\documentclass[runningheads]{llncs}

\usepackage{eccv}

\usepackage{eccvabbrv}

\usepackage{graphicx}
\usepackage{booktabs}

\usepackage{amsmath}
\usepackage{mathtools}
\usepackage{pifont}
\newcommand{\xmark}{\ding{55}}
\definecolor{LightRed}{rgb}{0.96,0.92,0.92}
\definecolor{LightGrey}{rgb}{0.9,0.9,0.9}
\usepackage[pagebackref,breaklinks,colorlinks,citecolor=eccvblue]{hyperref}
\newcommand{\pr}{P\&R}
\usepackage[table,svgnames]{xcolor}
\usepackage{colortbl}
\usepackage{bbm}
\usepackage{bbm}
\usepackage{amssymb}
\usepackage{multirow}
\usepackage{wrapfig}
\usepackage{enumitem}

\usepackage[accsupp]{axessibility}  

\usepackage{orcidlink}

\begin{document}

\title{Prime and Reach: Synthesising Body Motion for Gaze-Primed Object Reach}

\titlerunning{Prime and Reach}

\author{
Masashi Hatano\inst{1}$^{*}$ \and
Saptarshi Sinha\inst{2}$^{*}$ \and
Jacob Chalk\inst{2} \and
Wei-Hong Li\inst{2} \and \\ Hideo Saito\inst{1} \and Dima Damen\inst{2}\\}

\authorrunning{M.~Hatano et al.}

\institute{Keio University, Japan \and University of Bristol, UK\\
$^*$: Equal Contribution\\
\url{https://masashi-hatano.github.io/prime-and-reach/}\\}

\maketitle

\begin{centering}
\begin{minipage}{0.49\linewidth}
    \centering
    \includegraphics[width=\linewidth]{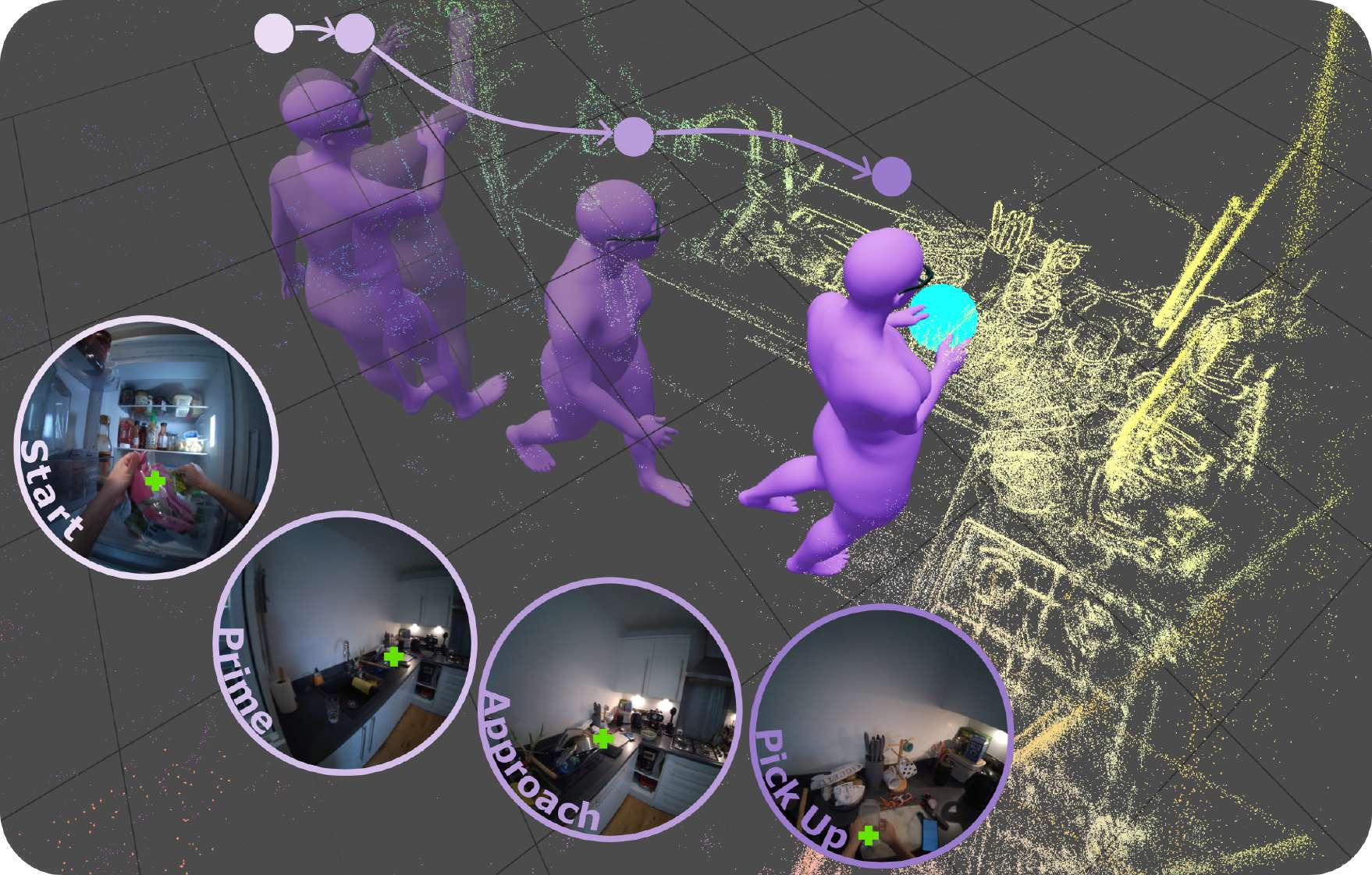}
\end{minipage}
\begin{minipage}{0.49\linewidth}
    \centering
    \includegraphics[width=\linewidth]{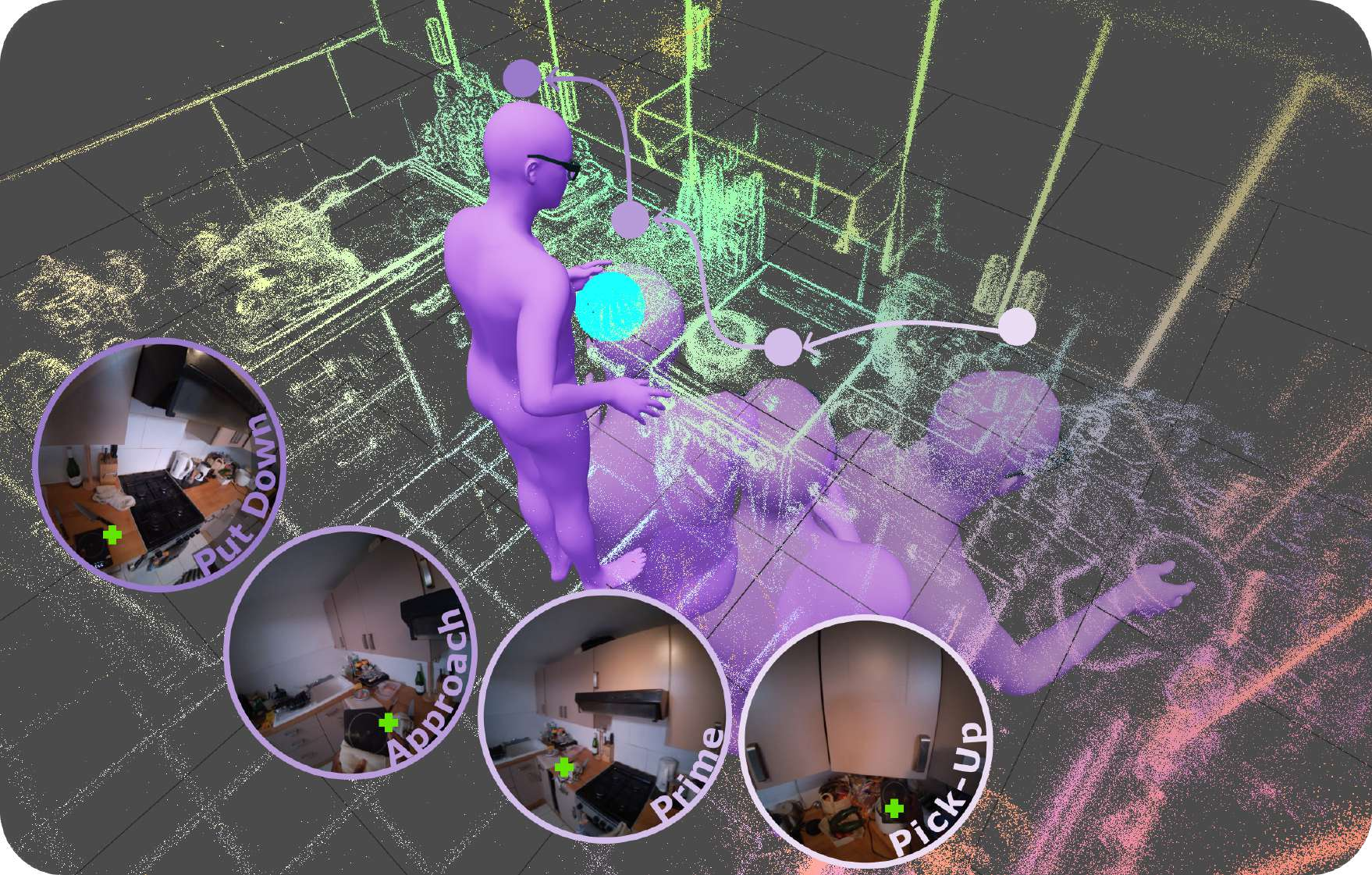}
\end{minipage}
\captionof{figure}{Prime \& Reach sequences from HD-EPIC~\cite{perrett2025HD-EPIC}, using full-body pose from EgoAllo~\cite{yi2025egoallo}.
\textbf{(Left)} A sequence starting with the intention to reach the container (\textcolor{cyan}{cyan sphere}). Gaze priming is evident -gaze (\textcolor{DarkGreen}{green plus}) intersecting the object- during the approach before reaching the object.
\textbf{(Right)} Similar behaviour is noted for priming and picking up the scale (\textcolor{cyan}{cyan sphere}). [darker colors indicate later time].}
\label{fig:egoallo_examples}
\end{centering}

\begin{abstract}
Human motion generation is a challenging task that aims to create realistic motion imitating natural human behaviour. 
We focus on the well-studied behaviour of priming an object/location for pick up or put down -- that is, the spotting of an object/location from a distance, known as gaze priming, followed by the motion of approaching and reaching the target location.
To that end, we curate, for the first time, 23.7K gaze-primed human motion sequences for reaching target object locations from five publicly available datasets, \ie, HD-EPIC, MoGaze, HOT3D, ADT, and GIMO.
We pre-train a text-conditioned diffusion-based motion generation model, then fine-tune it conditioned on goal pose or location, on our curated sequences.
Importantly, we evaluate the ability of the generated motion to imitate natural human movement through several metrics, including the `Reach Success' and a newly introduced `Prime Success' metric. 
Tested on 5 datasets, our model generates diverse full-body
motion, exhibiting both priming and reaching behaviour, and outperforming baselines and recent methods.
\end{abstract}
\section{Introduction}
Text-guided human motion generation works excel at translating textual descriptions into a wide array of human movements~\cite{tevet2023mdm, guo2023momask, Guo2022t2m}, from simple actions like running to complex motions such as dancing.
The scope has expanded to include navigating environments~\cite{diomataris2024wandr, karunratanakul2023gmd, xie2024omnicontrol} and interactions with objects~\cite{taheri2022goal, paschalidis20253d, tendulkar2023flex, xu2023interdiff, diller2024cg}.
However, these works heavily rely on synthetic datasets~\cite{tendulkar2023flex, araujo2023circle} or controlled in lab recordings~\cite{taheri2020grab, jiang2023full, lv2024himo}.
Consequently, they fail to model natural behaviours like prime and reach (\pr), limiting their utility as digital human replicas.

On the other hand, egocentric data collection provides a compelling alternative solution, enabling the capture of diverse daily activities~\cite{ma2024nymeria, grauman2024ego} and human-human interactions~\cite{zhang2022egobody} accompanied by a head-mounted eye-tracking camera.
Nevertheless, most egocentric datasets~\cite{Damen2022RESCALING, perrett2025HD-EPIC, grauman2022ego4d} do not capture full-body motion, as pairing egocentric and 3D sensors is costly and challenging.
Several recent works~\cite{li2023ego, yi2025egoallo, guzov-jiang2025hmd2, chi2025estimating} have explored estimating full-body motion conditioned on the egocentric camera pose and viewpoint, training on diverse data~\cite{ma2024nymeria, grauman2024ego}.
Egocentric datasets have not been explored for generative modelling, including the preparatory priming and reaching motion (see Fig~\ref{fig:egoallo_examples} for sample sequences).

Human visuomotor coordination is fundamentally anticipatory, continuously integrating sensory information to facilitate fluid, goal-directed actions~\cite{jia2025learning}. 
The anticipatory mechanism heavily relies on gaze. 
By directing visual attention to relevant objects or areas before reaching, gaze offers crucial predictive signals that enable the motor system to prepare and execute actions efficiently and naturally~\cite{land1999roles, johansson2001eye, hayhoe2005eye, jia2025learning}. 
The ability to prime and reach has recently been investigated in various real-world applications, including robotics~\cite{kerrj2025eyerobot,rahrakhshan2022learning,shafti2019gaze}. 
For example, in a manipulation involving a towel, the robot would visually fixate on the towel and the grasp point before extending its arm~\cite{kerrj2025eyerobot}.
However, such behaviour has not been explored in training or evaluating motion generation.

In this work:

\begin{itemize}[leftmargin=*,noitemsep, nolistsep,label=\textbullet]
    \item We enable, for the first time, the synthesis (or generation) of priming and reaching behaviour by combining egocentric datasets that offer gaze priming and full-body motion with conditioned diffusion models.
    \item We curate 23.7K prime and reach sequences from five datasets, by leveraging gaze and object-location annotations to automatically extract motion segments that contain priming and reaching.
    \item We design goal-conditioned models generating prime and reach motions, introducing a ‘Prime Success’ metric to evaluate against existing baselines.
    \item Using goal pose as a condition, our model can boost priming ability by up to $18.2\%$ absolute gain over the previous best-performing method, while achieving reach success nearly perfectly. When conditioning on the object location, our model improves the priming ability by up to $19.5\%$.
\end{itemize}
\section{Related Work}

Human motion generation aims to create realistic, continuous human movements that simulate or animate natural human motion. The majority of work in this field generates human motion from a single modality such as text~\cite{Guo2022t2m, petrovich22temos, petrovich23tmr, zhong2023attt2m, athanasiou2022teach}, action~\cite{guo2020action2motion, cervantes2022implicitneuralrepresentationsvariable, petrovich21actor, posegpt}, speech~\cite{zhu2023taming, alexanderson2023listen}, and music~\cite{li2021learn, tseng2022edge, sun2022nonfreezing}, with a few recent works tackling motion generation from multiple modalities~\cite{li2024unimotion, genmo2025, bian2024motioncraft, luo2024m3gpt}.

\vspace{0.05cm}
\noindent \textbf{Text-to-motion Generation.}
Initial efforts~\cite{ahuja2019language2pose, ghosh2021synthesis} in text-to-motion were deterministic, converging to an averaged motion given an input text.
After the advent of the denoising diffusion models~\cite{ho2020denoising, song2021denoising}, these models are nowadays a common practice to generate text-conditioned human motion~\cite{tevet2023mdm, zhang2024motiondiffuse, zhang2023remodiffuse, kim2023flame}.
MDM~\cite{tevet2023mdm}, MotionDiffuse~\cite{zhang2024motiondiffuse}, and FLAME~\cite{kim2023flame} are conditioned on features extracted by a pre-trained text encoder.
Another common approach is to disentangle motion representation from generation~\cite{guo2023momask, jiang2024motiongpt, chuan2022tm2t, zhang2023t2mgpt}. This is a two-stage process: first, a VQ-VAE~\cite{van2017neural} is trained to create a discrete codebook that tokenises continuous motion sequences. Then, a Transformer-based autoregressive model is trained on these discrete tokens to learn motion primitives by next-token-prediction.
Although the text can serve as a strong signal for conditioning semantic motion, these methods often lack precise control over body positions.

\vspace{0.05cm}
\noindent \textbf{Location-Conditioned Motion Generation.} 
Our work is related to a line of research that generates controllable human motion~\cite{pinyoanuntapong2025maskcontrol,xie2024omnicontrol, karunratanakul2024optimizing,tevet2025closd,karunratanakul2023gmd,Zhao:DartControl:2025,diomataris2024wandr}.
Guided Motion Diffusion (GMD)~\cite{karunratanakul2023gmd} extends text-to-motion models with spatial controls, including root trajectories, keyframe locations, obstacle avoidance, and sparse joint constraints.
DNO~\cite{karunratanakul2024optimizing} instead treats a pre-trained text-to-motion diffusion model as a motion prior and optimises the initial noise vector with task-specific gradients (\eg, goal joints or target location) at inference time, enabling flexible control without retraining.
Furthermore, DartControl~\cite{Zhao:DartControl:2025} proposed autoregressive prediction with latent motion diffusion conditioned on history sequences and textual description. It achieves motion-in-between via DNO and goal-reaching task through policy control.
Recently, WANDR~\cite{diomataris2024wandr} introduced a data-driven model conditioned on the initial pose and the goal location of the right wrist to generate avatars that walk and reach the goal in 3D space.
Other works~\cite{li2024egogen, diomataris2025movinglookingvisiondrivenavatar} address the goal-reaching human motion generation via egocentric perception and reinforcement learning.
Despite these advancements, these methods rely on synthetic datasets~\cite{araujo2023circle} or MoCap-based datasets~\cite{mahmood2019amass}, which limit their ability to generate natural interactions in real-world scenarios.
Additionally, these works do not address or evaluate priming.
In this work, we curate the first set of datasets that include full-body, priming, and reaching, with a focus on replicating this human priming-then-reaching motion through generation.

\vspace{0.05cm}
\noindent \textbf{Ego-body Pose Generation.} 
Recent research has explored estimating~\cite{jiang2022avatarposer, jiang2024egoposer, Luo2021DynamicsRegulatedKP, chi2025estimating, guzov-jiang2025hmd2} or forecasting~\cite{yuan2019ego, escobar2025egocast, patel25uniegomotion, Hatano2025EgoH4} human motion from an egocentric perspective.
These methods typically adopt a generative approach as the human body is largely invisible from an egocentric view, unlike ego-body pose estimation from a downward-facing camera~\cite{Millerdurai_EventEgo3D_2024, wang2024egocentric, hakada2022unrealego, tome2019xr, cuevas2024simpleego, zhao2021egoglass}.
EgoEgo~\cite{li2023ego} is the first work to propose head pose (camera pose) conditioned human motion generation, but was mainly evaluated on synthetic datasets.
Subsequently, EgoAllo~\cite{yi2025egoallo} proposes a head-centric representation (\ie, canonicalisation) to achieve spatial and temporal invariance, and also enables the integration of in-view 3D hand poses for better prediction.
We utilise this method to generate human motion on gaze-primed and reach sequences curated from egocentric datasets. 

\noindent \textbf{Eye-gaze in Motion.}
Eye-gaze is an important predictive signal that directs attention and primes the processing of future movements~\cite{land1999roles, johansson2001eye, hayhoe2005eye}. Recognising the critical role of gaze, recent research has been focusing on estimating gaze/saliency~\cite{lai2022eye, lai2023eye, chong2018connecting} or explicitly leveraging this cue for various problems, such as video understanding~\cite{mazzamuto2025gazing, peng2025eyemllmbenchmarkingegocentric} or human-robot interactions~\cite{stolzenwald2018can, saran2018human}. 
Several works focus on future motion prediction following gaze priming~\cite{wei2018and,yan2023gazemodiff,hu2025hoigaze,lou2024multimodal,jia2025learning}. Tian \etal~\cite{tian2024gaze} generate hand-object interactions but only in table-top settings. 
Different from these works, we wish to synthesise both the gaze priming and the reach motion, for the full body, conditioned on the goal.

\section{Prime and Reach Data Curation}

We first introduce the principle of curating `Prime and Reach' (\pr) sequences from longer videos (\cref{sec:sequence_curation}).
We then detail the steps we carried out to curate these sequences from five public datasets (\cref{sec: datasets}). We note the statistics of these sequences, which we use for training and evaluation.

\subsection{P\&R sequence Curation}
\label{sec:sequence_curation}
Interaction datasets include multiple and frequent object reaching and manipulation behaviours. However, a critical aspect largely unexplored is the role of gaze in priming or ``spotting'' objects prior to the reaching motion. 
We take this missed opportunity and curate for the first time \pr\ motion sequences from datasets capturing wearable gaze and object interactions.
We are inspired by the ``\textit{gaze priming}'' discussed in~\cite{perrett2025HD-EPIC} where objects were annotated in 3D and then used to identify the fixation that occurs prior to the physical action, signalling intent of interaction. 

Starting from long videos, we extract timestamps for object pick-up or put-down events. 
We note that priming takes place also during put-down where the future location of an object is primed before the action.
Given the known pick/put event at time $t_e$, we analyse a temporal window of duration $w$
immediately preceding it to find a moment $t_p \in [t_e - w, t_e]$.
We wish to identify when the user's gaze first attends to or primes the relevant location for the pick-up/put-down event.
For pick-up events, we associate the event with 3D location of the object. This location will be used to identify the priming event. Importantly, for put-down events, we instead use the future 3D location of the object (which at the start of the motion is an empty part of the 3D space) to search for the priming -- \ie, we track the intersection of the gaze of the camera wearer with this empty space, priming the location where the object is going to be placed.  

Specifically, we project the user's gaze into the 3D environment to form a ray. First, the gaze direction provided by the eye-tracker in the camera's local coordinate system at any time $t$, $\mathbf{p}^t_\text{gaze\_cam}$, is transformed using the camera-to-world transformation matrix $\mathbf{T}^t_\text{c2w}$. The final normalised gaze direction vector, $\hat{\mathbf{d}}^t_\text{gaze}$, is then computed as the vector from the camera's world position, $\mathbf{o}^{t}_\text{cam}$, to this new world-space gaze point $\mathbf{p}^{t}_\text{gaze\_world}$ as shown in Equation~\ref{eq:gaze_proj}.
\begin{gather}
\mathbf{p}^t_\text{gaze\_world} = (\mathbf{T}^t_\text{c2w} \mathbf{p}^t_\text{gaze\_cam}) \nonumber \\ 
\hat{\mathbf{d}}^t_\text{gaze} = \frac{\mathbf{p}^t_\text{gaze\_world} - \mathbf{o}^t_\text{cam}}{||\mathbf{p}^t_\text{gaze\_world} - \mathbf{o}^t_\text{cam}||}
\label{eq:gaze_proj}
\end{gather}
We register a relevant location as primed if the gaze ray, originating from the camera's position, intersects with the corresponding 3D bounding box or 3D location $o_{\text{3D}}$.
Therefore, we define the prime time $t_p$ as 
\begin{gather}
    T_{\text{int}} =\{t | t \in [t_e - w, t_e],\mathbb{I}(\text{intersect}(\hat{\mathbf{d}}^t_\text{gaze}, o_{\text{3D}})) = 1\} \nonumber \\
    t_p  = \min_{t \in T_{\text{int}}} t,
\label{eq:prime_time}
\end{gather}
where $T_{\text{int}}$ is the set of all timestamps within the temporal window $[t_e - w, t_e]$, where the gaze ray intersects the 3D location and $t_p$ is the first moment where the intersection happens. 
We discard sequences where $T_{\text{int}} = \emptyset$. 
Following \cite{perrett2025HD-EPIC}, we use $w=10$ secs.
To compute the intersection \ie $\text{intersect}(\hat{\mathbf{d}}^t_\text{gaze}, o_{\text{3D}})$, we use the slab test method~\cite{Majercik2018Voxel}, details of which are available in the supplementary.
At the end of this process, we get \pr\ sequences each defined by a prime time $t_p$ and reach time $t_e$.

\begin{table*}[t]
\centering
\caption{\textbf{Curated Dataset Statistics}. We report statistics on curated \pr\ sequences across five publicly available datasets, ordering them by the size of curated sequences. We report the number of \pr\ sequences, duration between prime time and reach time \ie $t_e - t_p$ (Prime Gap), body pose type, the distance/movement of body and hand. * indicates that body poses are estimated using~\cite{yi2025egoallo}.}
\centering
\resizebox{0.8\linewidth}{!}{
\begin{tabular}{l cccc cc }
\toprule
Dataset &
 
\#\pr\ Seq. & \begin{tabular}{c}Sequence \\ Duration ($s$)\end{tabular} & \begin{tabular}{c}Prime  \\ Gap ($s$) \end{tabular} & Body Pose Type  & \begin{tabular}{c}Body  \\ Movement ($m$) \end{tabular} & \begin{tabular}{c}Hand \\ Movement ($m$) \end{tabular} \\
\toprule
HD-EPIC~\cite{perrett2025HD-EPIC} & 18,134&$5.49 \pm 2.76$ & $3.55 \pm 2.79$& SMPL-H*~\cite{SMPL:2015}& $0.72\pm 0.67$  &$0.45\pm 0.22$ \\
MoGaze~\cite{kratzer2020mogaze} &  2,637& $3.64 \pm 0.94$ &$1.53 \pm 0.92$ &3D Skeleton & $1.07\pm 0.62$ & $0.75 \pm 0.25$ \\
HOT3D~\cite{banerjee2025hot3d} & 2,416& $4.31 \pm 1.54$ &$2.37 \pm 1.57$&SMPL-H*~\cite{SMPL:2015}& $0.20 \pm 0.15$  & $0.38 \pm 0.19$ \\
ADT~\cite{pan2023aria} & 411& $7.44 \pm 2.47$ &$4.51 \pm 2.74$ & 3D Skeleton & $1.23 \pm 1.28$& $0.56 \pm 0.24$\\
GIMO~\cite{zheng2022gimo} & 130& $7.11 \pm 2.49$ & $4.47 \pm 1.49$ & SMPL-X~\cite{pavlakos2019expressive} & $3.09 \pm 1.20$ & $0.64 \pm 0.19$\\
\bottomrule
\end{tabular}}
\vspace*{-12pt}
\label{table: stats curated dataset}
\end{table*}

\begin{figure}[t]
    \centering
    \includegraphics[width=\linewidth]{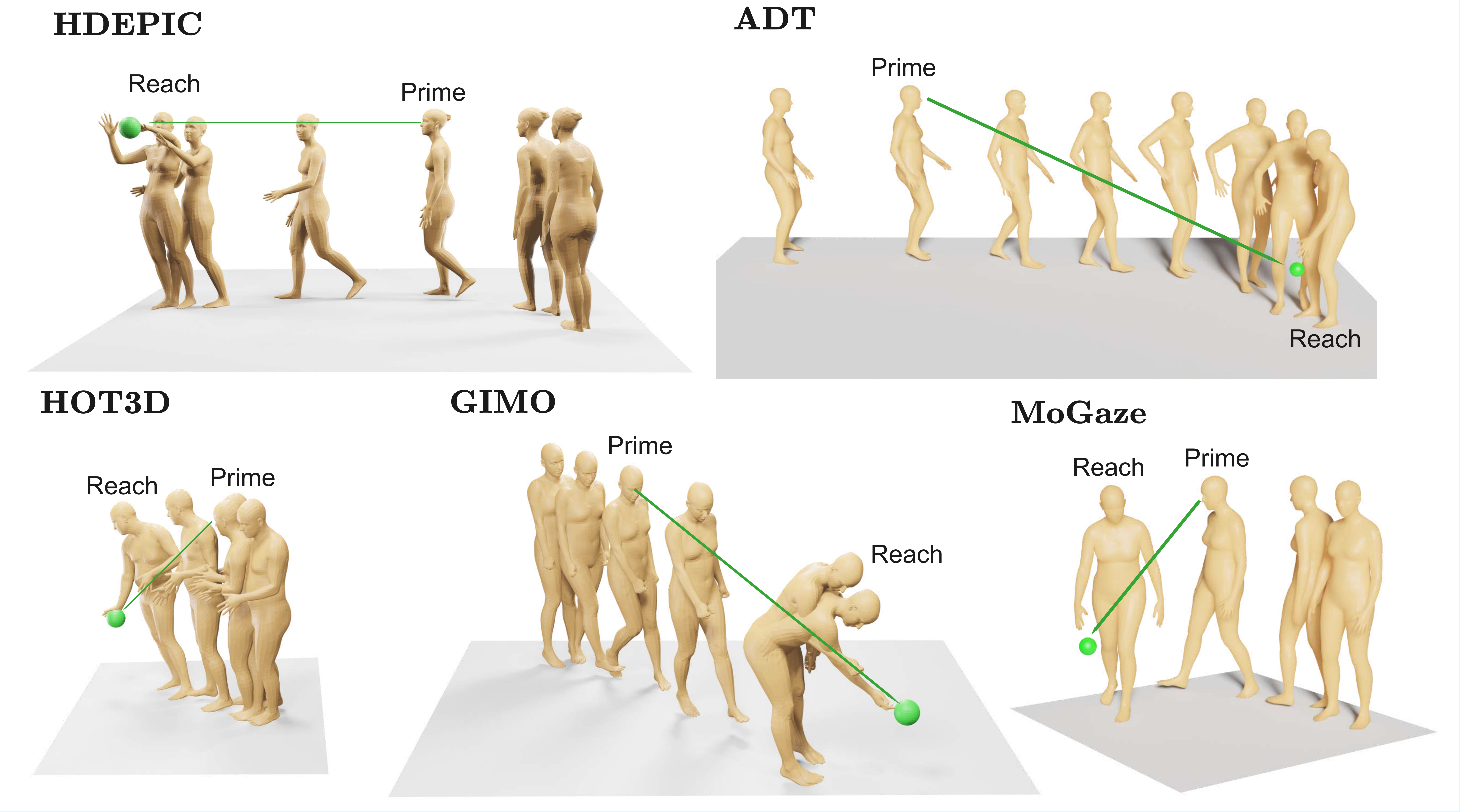}
    \vspace*{-18pt}
    \caption{Examples of curated \pr\ motion sequences from five different datasets.}
    \vspace*{-18pt}
    \label{fig:dataset_curation}
\end{figure}

\subsection{Datasets}
\label{sec: datasets}
As explained in \cref{sec:sequence_curation}, we formulate how \pr\ sequences can be curated from long video sequences. 
We consider five publicly available human-object interaction datasets, all of which contain 2D gaze tracked from wearable gaze trackers along with camera poses~\cite{perrett2025HD-EPIC,kratzer2020mogaze,banerjee2025hot3d,pan2023aria,zheng2022gimo}. 

\noindent \textbf{HD-EPIC}~\cite{perrett2025HD-EPIC} is an egocentric video dataset capturing diverse human-object interactions in the kitchen using the Aria Device~\cite{engel2023projectarianewtool}.
The dataset provides timestamp annotations of every object's pick/put events, along with 3D location and bounding boxes around the objects at the pick and put locations.
Using the 3D annotations with the gaze and camera pose in \cref{eq:gaze_proj,eq:prime_time}, we determine the priming time $t_p$ for each pick and put interaction in the dataset. 
We prepend these sequences by a fixed 2-second duration, so the start of the sequence is ahead of the priming, resulting in the sequence $[t_p-2$ secs$, t_e]$.
We use EgoAllo~\cite{yi2025egoallo} to estimate full-body motion for our \pr\ sequences as SMPL-H\cite{SMPL:2015} parameters.

\noindent \textbf{MoGaze}~\cite{kratzer2020mogaze} is human motion data designed explicitly for human-object interactions, with a particular focus on `pick' and `put'.
The dataset includes synchronised full-body motion captured using motion capture markers, 3D object models and their 6-DoF, and eye-gaze data.
The dataset contains 180 minutes of motion capture data from seven participants performing pick-and-put actions along with temporal segment annotations of these actions.
Using gaze and object locations, we determine priming timestamps ($t_p$) for each event. We slice the motion data for $t \in [t_p-2, t_e]$ constructing $2,637$ \pr\ sequences.

\noindent \textbf{HOT3D}~\cite{banerjee2025hot3d} captures 3D hand-object interactions.
The dataset offers 198 Aria recordings featuring 14 subjects interacting with 33 diverse objects.
We only use the Aria videos as these provide gaze information. 
As pick/put timestamp annotations are not provided,  
we extract temporal segments where an object is in-hand by thresholding ($< 5$ cm) the distance between the nearest hand vertices and the object locations.
After identifying these in-hand segments, we refine the segment boundaries by detecting the object state change from stationary to in-hand or vice-versa. This gives us pick/put events $t_e$.
We use gaze and camera pose to estimate the prime time for these events, resulting in $2,416$ 
\pr\ sequences. We estimate full-body motion for these sequences using EgoAllo~\cite{yi2025egoallo}.

\noindent \textbf{Aria Digital Twin (ADT)}~\cite{pan2023aria} provides a rich collection of synchronised data, including images, eye-tracking data, 6-DoF object data, and 3D human poses.
72 videos in the dataset capture indoor activities and interactions involving 398 unique objects and provide paired eye gaze and 3D body motion data.
We curate \pr\ sequences from these videos.
Same as HOT3D, we find temporal segments when objects are in-hand by thresholding the distance between object locations and nearest wrist locations from the corresponding body poses.
We identify pick and put events near the temporal boundaries of these segments based on how the object state changed. These events were then primed to determine $t_p$, resulting in 411 \pr\ sequences with full-body motion data.

\noindent \textbf{GIMO}~\cite{zheng2022gimo} is a benchmark that focuses on intent-guided human motion segments. It provides 217 trimmed segments, along with corresponding SMPL-X fitted IMU-captured body poses and egocentric views with eye gaze data captured by the HoloLens 2. We discard all segments that include resting activities without pick/put actions \eg, sitting or lying on the bed. GIMO segments include object pick-up but no put-down.
However, they do not provide the 3D locations or timestamps of object pick-up. Therefore, we manually annotate these event timestamps ($t_e$) from RGB videos and use the relevant wrist location at the timestamp as our object locations. We determine $t_p$ following the same method and get 130 \pr\ SMPL-X body motion sequences.

In total, we curate $23,728$ \pr\ sequences from the five datasets.
Statistics are provided in \cref{table: stats curated dataset} and sample \pr\ sequences in \cref{fig:dataset_curation}. Importantly, we unify body pose formats across datasets by representing them using the canonicalised 22-joint body motion used in HumanML3D~\cite{Guo_2022_CVPR}. Following \cite{tevet2023mdm}, we convert the 22 joint positions to 263-dim vector representation that combines local pose, rotation and velocity of each joint. 
The curated sequences are split 70\%-30\% into train and test sets.

\section{Method}
Here, we address the task of goal-conditioned human motion generation with the ability to prime and reach a given object.
Specifically, the task aims to generate human motion sequences $\{x^i\}_{i=1}^N$ of length $N$, where $x^i \in R^{J\times 3}$ represents 3D positions of $J$ body joints, guided either by desired \textbf{goal pose} or target \textbf{goal object location} as a condition.
We consider and compare the two conditions.

\subsection{Prime \& Reach Motion Diffusion Model}  \label{sec:prime_and_reach_method}
\begin{figure}[t]
    \centering
    \includegraphics[width=\linewidth]{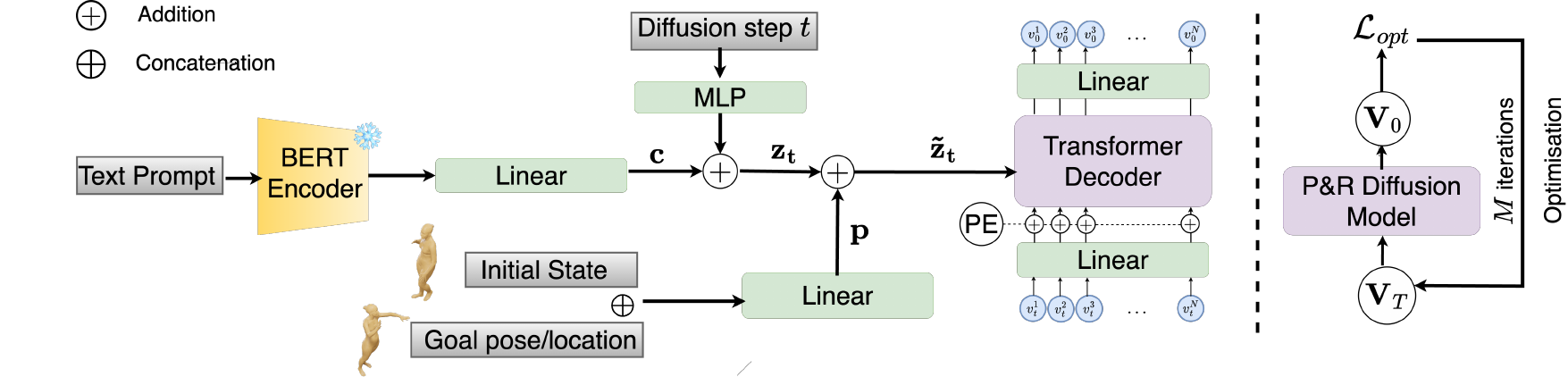}
     \vspace{-12pt}
    \caption{ \pr\ motion diffusion model for goal-conditioned motion generation. We concatenate the initial state of the human body and the goal pose/goal object as conditions, along with a text condition describing the type of action the motion is expected to perform. This accumulated condition is injected into the transformer decoder layers, which then outputs an $N$-length motion sequence over multiple diffusion steps. At inference we perform diffusion latent noise optimisation over $M$ iterations using the same conditioning (i.e. initial state and goal pose or location).}
    \vspace*{-12pt}
    \label{fig:prime_and_reach_model}
\end{figure}

\noindent \textbf{Conditioning.} Diffusion models have demonstrated exceptional capability for text-conditioned motion generation \cite{tevet2023mdm,zhang2023remodiffuse}.
Motivated by this, we use a diffusion generative model, as in~\cite{tevet2023mdm} for our task. We present our architecture in \cref{fig:prime_and_reach_model}. Starting from pure noise at $t=T$, the transformer decoder generates motion through iterative denoising over multiple diffusion timesteps $t=\{T, ..., 0\}$ where $t=0$ produces the predicted motion. This generation is guided through a set of conditions injected into the decoder:
\begin{itemize}[leftmargin=*,noitemsep, nolistsep,label=\textbullet]
    \item \textbf{Text prompt}: This allows the model to benefit from text-to-motion pre-training. We describe the action as \eg, `The person moves across and picks/puts an object.' We use the knowledge of the action (\ie, whether it is a pick up or a put down) in both training and inference to guide the synthesis. We refer to this conditioning text as $\mathbf{c}$.
    \item \textbf{Initial state} of the body describes where and how the motion initiates.
    As our \pr\ sequences do not start from a static or neutral pose, but are sampled from within a longer sequence, it is important to feed the initial pose and the velocity, as this impacts the guided motion.
    We represent this by the starting pose $(\hat{x}^1) \in R^{J\times3}$ and the joint velocities \ie $(\hat{x}^1 - \hat{x}^0)$ where $\hat{x}^0$ is the preceding frame before the curated \pr\ sequence. We concatenate the start pose and joint velocities as a single vector, which forms our initial state.
    \item \textbf{Goal.} In addition to the initial pose, motion generation expects the goal of the motion to be specified. We evaluate two possible goal formulations for \pr. The first is the final goal/target pose at the end of the motion $(\hat{x}^N)$. 
    The \textbf{goal pose} not only guides the motion to reach the object but additionally guides where the agent would stand relative to the object (through the full pose at the end of the motion) as well as which hand would be reaching the object (guided through the position of the hand joints).
    Second, we use the more challenging goal of only specifying the \textbf{object location} $(o_{\text{3D}} \in R^3)$ as a condition. Given only the object location as goal, the model has more freedom to generate motions choosing the relative body pose to the object.
\end{itemize} 
Importantly, we do not believe priming to be a condition. It should be implicitly learnt from the data. When using the \textbf{goal pose}, the direction where the object is, and thus how it can be primed is implicit. When using the goal object \textbf{location}, priming is synthesised by the body attending to this exact location.

\noindent \textbf{\pr\ motion generation with condition.} 
Next, we explain how we use these conditions for the generation process.
First, we encode the input text prompt $\mathbf{c}$ using a pre-trained text encoder~\cite{sanh2019distilbert}.
Both the diffusion noise time step $t$ and $\mathbf{c}$ are then projected to a latent space and summed to get a token $\mathbf{z_t}$. Note that for generating motion conditioned on text, $\mathbf{z_t}$ is directly injected into the transformer decoder's layers. We pre-train this model for text-conditioned motion generation. 
This pre-training enables the model to learn prior knowledge of fine-grained full-body motion involved in everyday activities.

For learning \pr\ motion generation, we initialise our model with the pre-trained text-to-motion model weights and fine-tune it with all three conditions. To add the initial state and goal condition, we first flatten and concatenate them into a single 1D vector. The resultant vector is linearly projected to a latent space to give $\mathbf{p}$.  $\mathbf{p}$ is injected through residual addition to the text condition $\mathbf{z_t}$ as $\mathbf{\tilde{z}_t} =\mathbf{z_t} + \mathbf{p}$.

This allows the additional conditions to modulate the global context without introducing new attention pathways (e.g., cross-attention) that could disrupt the pre-trained text-conditioned motion prior (ablated in supplementary).
This modified condition $\mathbf{\tilde{z}_t}$ is injected into the transformer decoder's layers through cross-attention blocks and guides the generation over multiple diffusion timesteps. 

Once de-noised, the decoder produces 263-dim representations of body joints $\{v^n\}_{n=1}^N$ where $v^n \in R^{263}$ combines local position, local rotation, and local velocity of all 22 body joints. This is post-processed as $\{x^n\}_{n=1}^N = g(\{v^n\}_{n=1}^N)$ to get the predicted 22 joint positions. We use the same joints-to-latent dimension conversion function $g$ as in \cite{karunratanakul2023gmd,tevet2023mdm}.

\subsection{Training and Inference of \pr\ Model}  
During training, following \cite{tevet2023mdm}, the model is optimised to reduce the reconstruction error between the 263-dim representations of generated and ground truth motion sequence \ie $\mathcal{L} = \sum_{n=1}^N ||\hat{v}^n - v^n||_2^2$, where $\{\hat{v}^n\}_{n=1}^{N}$ is the 263-dim representation from the ground truth motion $v^n = g^{-1}(\hat{x}^n)$. We add a joint reconstruction loss as $\mathcal{L}_{\text{joint}} = \sum_{n=1}^N ||\hat{x}^n - x^n||_2^2$ which acts on the original 22 joint pose. Our total loss is $\mathcal{L} + \mathcal{L}_{\text{joint}}$.

At inference, in addition to conditioning, we perform diffusion latent noise optimisation on the generated motion in line with prior works~\cite{Zhao:DartControl:2025,karunratanakul2024optimizing,duran2026fusionfullbodyunifiedmotion}.
We treat the diffusion noise $V_T$ as a
latent variable and generate motion with the full sampling
process of the trained \pr\ model, with gradients propagating through
all $T$ denoising steps.
We use the same conditioning during this optimisation step (\ie initial and goal pose or initial and goal condition).
We consider $\mathcal{L}_{\text{init}} = ||\hat{x}^1 - x^1||_2^2$ is the mean squared error between predicted and ground truth starting poses.
For goal-pose conditioned generation, we use $\mathcal{L}_{\text{goal}} = ||\hat{x}^N - x^N||_2^2$.
For target location conditioning, we design an objective for the motion to reach the object location $o_{\text{3D}}$ with the right wrist \ie $\mathcal{L}_{\text{goal}} = ||o_{\text{3D}} - x^N_{\text{right wrist}}||_2^2$. 

We then calculate the optimisation loss, $\mathcal{L}_{\text{opt}} = \mathcal{L}_{\text{init}} + \mathcal{L}_{\text{goal}} + \alpha \mathcal{L}_{\text{jerk}}$ where $\mathcal{L}_{\text{jerk}}$ controls the quality of the motion following \cite{Zhao:DartControl:2025}.
This optimisation is iteratively performed over $M$ iterations (right figure in \cref{fig:prime_and_reach_model}).

\section{Experiments}
\label{sec:experiments}
Here we explain implementation details~(\cref{sec:impl}), evaluation metrics~(\cref{sec:eval_metrics}), baselines (\cref{sec:comparison}), experimental results (\cref{sec:quantitative}) and ablations (\cref{sec:ablation}).

\subsection{Implementation Details}
\label{sec:impl}
We pre-train our \pr\ motion diffusion model for text-conditioned motion generation on the large-scale Nymeria \cite{ma2024nymeria} dataset to learn the motion prior of everyday activities. 
Nymeria provides large-scale full-body motion data of participants performing diverse actions, including some priming and reaching activities, captured by Xsens mocap sensors, accompanied by atomic narrations describing the actions. 
This makes it a better pre-training dataset for \pr\ compared to the alternative HumanML3D~\cite{Guo_2022_CVPR}.
The narrations are used as text guidance. For pre-training, an initial learning rate of  $1e-4$ is used, for a maximum of 600K steps. 
We use a motion length of $N=150$ and classifier-free guidance with a probability of 0.2.
The pre-training takes $\sim36$ hours on one H200 GPU.

Initialised with the pre-trained weights, we fine-tune our \pr\ model on the training split of our curated \pr\ dataset, training a single model on 16.7K \pr\ sequences.
For fine-tuning, we use a learning rate of $5e-5$ for 250K steps, which takes approximately 15-20 hours.
We use $T=50$ diffusion steps in pre-training, fine-tuning, and inference, following~\cite{tevet2023mdm}.
During inference, we optimise the latent noise over $M=400$ iterations with an initial learning rate of 0.05 during $\mathcal{L}_{\text{opt}}.$ Following \cite{duran2026fusionfullbodyunifiedmotion}, we use $\alpha=0.5$.

\subsection{Evaluation Metrics} \label{sec:eval_metrics}
We report results on six metrics: two to directly evaluate our ability to prime and reach, two to evaluate the body pose at the goal, and two to evaluate the entire generated motion.

\noindent \textbf{(1) Prime Success}. This is evaluating whether the generated motion is priming the target location before reaching it. Specifically, within a \textit{generous} temporal window around the ground truth priming time $t_p$, we evaluate if the predicted motion exhibits intentional priming behaviour. 
While the generated motion can prime the object at a different time than the ground-truth, priming should happen well before reaching, not too far from the priming time.
The intentional priming behaviour requires moving one's gaze to look at the object, then attending to it.
We formulate this as follows:
 1) the head forward vector of the prediction motion ($\mathbf{\hat{H}}$) intersects the target location $o_{\text{3D}}$ within a proximity threshold for a minimum duration, and 2) the angular velocity of the head forward vector decelerates into the target location, signifying intentional priming from elsewhere. The reach metric per sequence is calculated as follows:
\begin{equation}
\mathbb{I} \{ \exists t \in [t_p - \sigma, t_p + \sigma] \mid \big( \forall k \in [t, t+\tau], dist(\mathbf{\hat{H}}_k, o_{\text{3D}}) \leq \delta \big) \land \Delta{\omega}_t < 0 \},
\end{equation}
where $\mathbb{I}$ is the indicator function and $dist(:,:)$ computes the orthogonal distance between $o_{\text{3D}}$ and the head forward vector. The term $\Delta{\omega}_t$ represents the change in angular velocity of the head forward vector at the onset of intersection; a negative value ($\Delta{\omega}_t < 0$) indicates the active deceleration toward the target. We use $\tau=0.1$ sec, $\delta=25$ cm, and $\sigma=1.0$ sec. Having $\sigma$ relaxes the metric to allow predicted motion to prime the object in a temporal window of $2\sigma$ around $t_p$.
The prime success then considers the percentage of sequences where an object is deemed primed.

\noindent \textbf{(2) Reach Success}. Following \cite{diomataris2024wandr}, this is the percentage of predicted sequences where either wrist reach within 10 cm of the goal.

\noindent \textbf{(3) Location Error}. 
Following \cite{karunratanakul2023gmd}, this is the percentage of motions, where final pelvis location of prediction is $\geq$ 50 cm away from ground truth pelvis.

\noindent \textbf{(4) Goal MPJPE}, which calculates the error between the final full body pose of the ground truth and the predicted motion. This includes the error in all joints including the hand reaching the object. Notice that this error assumes the same hand is reaching out to the object as the ground truth.

\noindent \textbf{(5) Mean Per Joint Position Error (MPJPE)}, which averages the joint position error in Euclidean distance over all generation frames.

\noindent \textbf{(6) Foot Skating}, a common evaluation of the generated human motion~\cite{karunratanakul2023gmd}, which measures the proportion of frames where foot slides while maintaining contact with the ground.

\begin{table*}[!htp]
\caption{\textbf{Comparison of motion generation baselines} on our curated \pr\ sequences using different metrics. While we train a single model for all datasets, we separate results per dataset. We show results for test splits of HD-EPIC, MoGaze, HOT3D, ADT, and GIMO separately. The baselines are grouped by the type of conditioning used for generation. $\dagger$ denotes the zero-shot inference. For MDM, we evaluate two pre-trained models: (1) trained on HumanML3D~\cite{Guo_2022_CVPR}, and (2) trained on Nymeria.\cite{ma2024nymeria}, denoted as $\ddagger$ and $^*$, respectively. Entries without a marker correspond to models fine-tuned on our \pr\ sequences.}
\label{table:baseline_comparison}
\begin{subtable}[t]{\linewidth}
\resizebox{\linewidth}{!}{
\begin{tabular}{cccccccc|cccccc}
\toprule
& & \multicolumn{6}{c|}{HD-EPIC}& \multicolumn{6}{c}{MoGaze}\\
\midrule
Condition& Method& \begin{tabular}[c]{@{}c@{}}Prime \\ Success\end{tabular} $\uparrow$ & \begin{tabular}[c]{@{}c@{}}Reach \\ Success\end{tabular} $\uparrow$ & \begin{tabular}[c]{@{}c@{}}Goal \\ MPJPE\end{tabular} $\downarrow$  & Loc Err $\downarrow$ &  MPJPE $\downarrow$ & \begin{tabular}[c]{@{}c@{}}Foot \\ Skating\end{tabular} $\downarrow$ & \begin{tabular}[c]{@{}c@{}}Prime \\ Success\end{tabular} $\uparrow$&\begin{tabular}[c]{@{}c@{}}Reach \\ Success\end{tabular} $\uparrow$ & \begin{tabular}[c]{@{}c@{}}Goal \\ MPJPE\end{tabular} $\downarrow$    & Loc Err $\downarrow$  & MPJPE $\downarrow$ & \begin{tabular}[c]{@{}c@{}}Foot \\ Skating\end{tabular} $\downarrow$ \\
\midrule
No condition & Static & 0.00 & 23.16 & 0.70 & 50.91 & 0.45 & \multicolumn{1}{c|}{--} & 0.00 & 2.56 & 1.06 & 75.99 & 0.62 & -- \\
\midrule
\multirow{3}{*}{Text} & MDM $\dagger \ddagger$ &9.53& 13.94 &1.14&81.75 &0.96 & \multicolumn{1}{c|}{0.16} & 2.85 & 2.20 & 2.03 & 94.11 & 1.45 & 0.39 \\
& MDM $\dagger ^*$  & 9.52& 13.47 &0.85&62.13 &0.59 &\multicolumn{1}{c|}{0.06} & 3.11 & 1.85 & 1.19 & 79.76 & 0.73 & 0.06 \\
& MDM & 12.40 &  18.76  & 0.84 & 58.73& 0.54 & \multicolumn{1}{c|}{0.04} & 5.27 & 2.90 & 1.24 & 85.42 & 0.72 & 0.05 \\
\midrule
\multirow{6}{*}{\begin{tabular}[l]{@{}l@{}}+ Initial State \\ \& Goal Pose\end{tabular}}
& GMD~$\dagger$\cite{karunratanakul2023gmd} & 33.27 & 30.77 & 0.26 & 2.00 & 0.32& \multicolumn{1}{c|}{0.10} & 8.11 & 10.86 & 0.35 & 2.23 & 0.54 & 0.11 \\
& GMD\cite{karunratanakul2023gmd} & 39.47 & 32.36  & 0.27 & 5.20 & 0.31& \multicolumn{1}{c|}{0.06} & 22.11 & 21.66 & 0.30 & 1.15 & 0.55 & 0.29 \\
& DNO~$\dagger$\cite{karunratanakul2024optimizing} & 43.42 & 87.44 & 0.07 & 2.60 & 0.30 & \multicolumn{1}{c|}{0.04} & 17.39 & 69.71 & 0.09 & 6.81 & 0.64 & \textbf{0.07} \\
& DNO~\cite{karunratanakul2024optimizing} & 48.30 & 87.44 &\textbf{0.05}  & \textbf{0.51} & 0.27 & \multicolumn{1}{c|}{0.07} & 29.27 & 34.21 & 0.09 & 0.55 & 0.62 & 0.16 \\
& DartControl~$\dagger$~\cite{Zhao:DartControl:2025} & 30.06 & 81.23 & 0.14 & 2.20 & 0.43 & \multicolumn{1}{c|}{0.15} & 38.26 & 71.16 & 0.11 & 0.14 & 0.52 & 0.35 \\
& DartControl~\cite{Zhao:DartControl:2025} & 35.29 & 88.27 & 0.11 & 1.86 & 0.38 & \multicolumn{1}{c|}{0.09} & 35.07 & 62.61 & 0.12 & 0.14 & 0.60 & 0.51\\
\rowcolor{LightRed}
\cellcolor{white} & P\&R & \textbf{52.75} & \textbf{97.40} & 0.09 & 1.58&\textbf{0.21} & \multicolumn{1}{c|}{\textbf{0.03}} & \textbf{45.24} & \textbf{96.47} &\textbf{0.05} & \textbf{0.00} & \textbf{0.32} & 0.16 \\
\midrule
\multirow{6}{*}{\begin{tabular}[l]{@{}l@{}}+ Initial State \\ \& Object Loc.\end{tabular}}
& WANDR~$\dagger$\cite{diomataris2024wandr}& 33.28 & 80.92 & 0.47 &40.35 &0.42 & \multicolumn{1}{c|}{0.11} & 31.74 & 96.81 & 0.62 & 60.14 & 0.61 & 0.25 \\
& WANDR~\cite{diomataris2024wandr} &31.92&75.16&0.54&47.65&0.50&\multicolumn{1}{c|}{0.16} &40.87&98.26&0.68&63.77&0.64&0.25 \\ 
& DNO~$\dagger$\cite{karunratanakul2024optimizing} & 37.34 & \textbf{100.00} & 0.47  & 37.87 & 0.44 & \multicolumn{1}{c|}{0.05} & 27.24 & \textbf{100.00}& 0.64 & 59.56 & 0.87 & \textbf{0.09} \\
& DNO~\cite{karunratanakul2024optimizing} & 45.42 & \textbf{100.00} & 0.41  & 34.64 & \textbf{0.27} & \multicolumn{1}{c|}{0.06} & 31.69 & \textbf{100.00} & 0.74 & 79.13 & 0.96 & 0.29 \\
& DartControl~$\dagger$~\cite{Zhao:DartControl:2025} & 28.82 & 89.20 & 0.47 & 40.82 & 0.52 & \multicolumn{1}{c|}{0.13} & 42.90 & \textbf{100.00} & 0.54 & 50.29 & 0.72 & 0.40\\
& DartControl~\cite{Zhao:DartControl:2025} & 30.04 & 89.42 & 0.44 & 34.16 & 0.47 & \multicolumn{1}{c|}{0.08} & 40.14 & \textbf{100.00} & 0.56 & 54.35 & 0.74 & 0.54\\
\rowcolor{LightRed}
\cellcolor{white} & P\&R & \textbf{51.00}& \textbf{100.00} & \textbf{0.38}&\textbf{25.26} & \textbf{0.27}& \multicolumn{1}{c|}{\textbf{0.03}} & \textbf{62.37} & \textbf{100.00} & \textbf{0.46} & \textbf{36.43} & \textbf{0.56} & 0.11 \\
\bottomrule
\end{tabular}}
\end{subtable}
\begin{subtable}[t]{\linewidth}
\resizebox{\linewidth}{!}{
\begin{tabular}{cccccccc|cccccc}
\toprule
& & \multicolumn{6}{c|}{HOT3D}& \multicolumn{6}{c}{ADT}\\  
\midrule
Condition& Method & \begin{tabular}[c]{@{}c@{}}Prime \\ Success\end{tabular} $\uparrow$ & \begin{tabular}[c]{@{}c@{}}Reach \\ Success\end{tabular} $\uparrow$ &  \begin{tabular}[c]{@{}c@{}}Goal \\ MPJPE\end{tabular} $\downarrow$  & Loc Err $\downarrow$ & MPJPE $\downarrow$   & \begin{tabular}[c]{@{}c@{}}Foot \\ Skating\end{tabular} $\downarrow$ & \begin{tabular}[c]{@{}c@{}}Prime \\ Success\end{tabular} $\uparrow$ & \begin{tabular}[c]{@{}c@{}}Reach \\ Success\end{tabular} $\uparrow$ & \begin{tabular}[c]{@{}c@{}}Goal \\ MPJPE\end{tabular} $\downarrow$  & Loc Err $\downarrow$ & MPJPE $\downarrow$ & \begin{tabular}[c]{@{}c@{}}Foot \\ Skating\end{tabular} $\downarrow$\\
\midrule
No condition & Static & 0.00 & 26.43& 0.35 & 22.01 & 0.32 & \multicolumn{1}{c|}{--} & 0.00 & 9.90 & 1.86 & 69.27 & 1.03 & -- \\
\midrule
\multirow{3}{*}{Text} & MDM $\dagger \ddagger$ & 27.28 & 6.25 & 1.11 & 75.85 & 0.89 & \multicolumn{1}{c|}{0.31} & 9.38 & 5.21 & 2.54 & 97.40 & 1.76 & 0.35 \\
 & MDM $\dagger ^*$ & 8.65 & 4.30 & 0.51 & 36.20 & 0.44 & \multicolumn{1}{c|}{0.00} & 3.65 & 6.25 & 1.98 & 83.33 & 1.16 & 0.06 \\
 & MDM & 19.25 & 30.32 & 0.36 & 14.45 & 0.32& \multicolumn{1}{c|}{0.00} & 10.58 & 15.29 & 2.23 & 90.63 & 1.23 & 0.17 \\
\midrule
\multirow{6}{*}{\begin{tabular}[l]{@{}l@{}}+ Initial State \\ \& Goal Pose\end{tabular}}
& GMD~$\dagger$\cite{karunratanakul2023gmd} & 31.85 & 20.16 & 0.38 & 25.00 & 0.37 & \multicolumn{1}{c|}{0.02} & 16.47 & 12.94 & 0.35 & 10.58 & 0.48 & 0.21 \\
& GMD~\cite{karunratanakul2023gmd} & 44.08 & 33.19 & 0.37 & 8.87 & 0.38 & \multicolumn{1}{c|}{0.03} & 21.18 & 23.53 & 0.37 & 9.41 & 0.41 & 0.24 \\

& DNO~$\dagger$\cite{karunratanakul2024optimizing} & 45.83 & 90.18 & 0.05 & 9.67 & 0.23 & \multicolumn{1}{c|}{0.02} & 25.88 & 52.94 & 0.04 & 5.88 & 0.56 & \textbf{0.08} \\
& DNO~\cite{karunratanakul2024optimizing} & 54.30 & 93.41 & \textbf{0.04} & 1.15 & 0.19 & \multicolumn{1}{c|}{0.02} & 28.23 & 80.00 & 0.04 & 1.00 & 0.45 & 0.16 \\
& DartControl~$\dagger$~\cite{Zhao:DartControl:2025} & 40.51 & 89.77 & 0.17 & 2.15 & 0.24 & \multicolumn{1}{c|}{0.01} & 14.12 & 67.06 & 0.22 & 8.24 & 0.62 & 0.24 \\
& DartControl~\cite{Zhao:DartControl:2025} & 45.76 & 89.77 & 0.18 & 2.02 & 0.28 & \multicolumn{1}{c|}{0.02} & 20.00 & 56.47 & 0.25 & 10.59 & 0.66 & 0.29\\
\rowcolor{LightRed}
\cellcolor{white} & P\&R & \textbf{58.65} & \textbf{99.19} & \textbf{0.04} & \textbf{0.00} & \textbf{0.14} & \multicolumn{1}{c|}{\textbf{0.00}} & \textbf{35.42} & \textbf{98.82} & \textbf{0.02} & \textbf{0.00} & \textbf{0.37} & 0.12 \\
\midrule
\multirow{6}{*}{\begin{tabular}[l]{@{}l@{}}+ Initial State \\ \& Object Loc.\end{tabular}}
& WANDR~$\dagger$\cite{diomataris2024wandr} & 34.32 & 91.66 & 0.33 & 17.77 & \textbf{0.27} & \multicolumn{1}{c|}{0.05} & 7.65 & 82.94 & 0.66 & 61.76 & \textbf{0.60} & 0.22\\ 
& WANDR~\cite{diomataris2024wandr} &45.09&83.58&0.43&31.49&0.31& \multicolumn{1}{c|}{0.06} & 
30.59 &80.00&0.79&70.59&0.64&0.28\\ 
& DNO~$\dagger$~\cite{karunratanakul2024optimizing} & 46.77 & \textbf{100.00} & 0.41 & 40.86 & 0.36 & \multicolumn{1}{c|}{0.04} & 23.53 & \textbf{100.00} & 0.64 & 61.18 & 0.70 & \textbf{0.10} \\
& DNO~\cite{karunratanakul2024optimizing}& 63.04 & \textbf{100.00} & 0.35 & 22.17 & 0.31 & \multicolumn{1}{c|}{0.05} & 37.64 & \textbf{100.00} & 0.62 & 60.78 & 0.61 & \textbf{0.10} \\
& DartControl~$\dagger$~\cite{Zhao:DartControl:2025} & 46.43 & 97.58 & 0.38 & 19.78 & 0.34 & \multicolumn{1}{c|}{0.01} & 25.88 & 97.65 & 0.69 & 64.71 & 0.86 & 0.29 \\
& DartControl~\cite{Zhao:DartControl:2025} & 48.59 & 97.71 & \textbf{0.32} & \textbf{8.88} & 0.30 & \multicolumn{1}{c|}{\textbf{0.00}} & 20.00 & 96.47 & 0.68 & 62.35 & 0.79 & 0.32 \\
\rowcolor{LightRed}
\cellcolor{white} & P\&R & \textbf{68.32}& \textbf{100.00} & 0.40 & 11.90 & 0.29 & \multicolumn{1}{c|}{0.01} & \textbf{52.08} & \textbf{100.00} &\textbf{0.57} & \textbf{52.54} & 0.61 & 0.11\\

\bottomrule 
\end{tabular}}
\end{subtable}

\begin{minipage}[t]{0.59\linewidth}
\resizebox{\linewidth}{!}{
\begin{tabular}{cccccccc}
\toprule
& & \multicolumn{6}{c}{GIMO}\\  
\midrule
Condition& Method & \begin{tabular}[c]{@{}c@{}}Prime \\ Success\end{tabular} $\uparrow$ & \begin{tabular}[c]{@{}c@{}}Reach \\ Success\end{tabular} $\uparrow$ & \begin{tabular}[c]{@{}c@{}}Goal \\ MPJPE\end{tabular} $\downarrow$  & Loc Err $\downarrow$ & MPJPE $\downarrow$   & \begin{tabular}[c]{@{}c@{}}Foot \\ Skating\end{tabular} $\downarrow$ \\
\midrule
No condition & Static & 0.00 & 0.00 & 3.41 & 100.00 & 1.86 & -- \\
\midrule
\multirow{3}{*}{Text} & MDM $\dagger \ddagger$ & 0.00 & 0.00 & 3.69 & 100.00 & 2.17 & 0.45 \\
\multicolumn{1}{c}{} & MDM $\dagger ^*$ & 0.00  & 0.00 & 3.34 & 100.00 & 1.82 & 0.08 \\
\multicolumn{1}{c}{} & MDM & 0.00 & 0.00 & 2.96 & 100.00 & 1.64 & 0.14 \\ 
\midrule
\multirow{6}{*}{\begin{tabular}[l]{@{}l@{}}+ Initial State \\ \& Goal Pose\end{tabular}}
& GMD~$\dagger$\cite{karunratanakul2023gmd} & 13.63 & 4.76 & 0.44 & 11.90 & 0.62 & 0.43 \\
& GMD~\cite{karunratanakul2023gmd} & 27.27 & 9.09 & 0.45 & 4.54 & 0.61 & 0.25 \\
& DNO~$\dagger$~\cite{karunratanakul2024optimizing} & 18.18 & 27.27 & 0.15 & 9.09 & 0.80 & 0.18 \\
& DNO~\cite{karunratanakul2024optimizing} & 31.82 & 72.72 & 0.09 & 4.54 & 0.65 & 0.15 \\
& DartControl~$\dagger$~\cite{Zhao:DartControl:2025} & 9.52 & 14.29 & 0.33 & 9.52 & 1.37 & \textbf{0.07} \\
& DartControl~\cite{Zhao:DartControl:2025} & 19.05 & 14.29 & 0.32 & 9.52 & 0.84 & \textbf{0.07} \\
\rowcolor{LightRed}
\cellcolor{white} & P\&R & \textbf{50.00} & \textbf{90.90} & \textbf{0.03} & \textbf{0.00} & \textbf{0.51} & 0.10 \\
\midrule
\multirow{6}{*}{\begin{tabular}[l]{@{}l@{}}+ Initial State \\ \& Object Loc.\end{tabular}}
& WANDR~$\dagger$\cite{diomataris2024wandr} & 14.29 & 61.90 & 0.51 & 57.14 & 0.79 & 0.44 \\ 
& WANDR~\cite{diomataris2024wandr} & 9.52 &52.38&0.65&71.43&0.77&0.42\\ 
& DNO~$\dagger$~\cite{karunratanakul2024optimizing} & 27.27 & \textbf{100.00} & 0.76 & 63.63 & 0.98 & 0.11 \\
& DNO~\cite{karunratanakul2024optimizing} & \textbf{40.91} & \textbf{100.00} & 0.39 & \textbf{40.91} & 0.72 & 0.11 \\
& DartControl~$\dagger$~\cite{Zhao:DartControl:2025} & 4.76 & 76.19 & 0.49 & 61.90 & 1.51 & \textbf{0.10}\\
& DartControl~\cite{Zhao:DartControl:2025} & 19.05 & 76.19 & 0.55 & 80.95 & 0.98 & 0.15\\
\rowcolor{LightRed}
\cellcolor{white} & P\&R & \textbf{40.91} & \textbf{100.00} & \textbf{0.46} & \textbf{40.91} & \textbf{0.65} & 0.25 \\
\bottomrule 
\end{tabular}}

\end{minipage}
\hspace{1pt}
\begin{minipage}[t]{0.38\linewidth}
        \centering
        \vspace{-2cm}
        \includegraphics[width=0.99\linewidth]{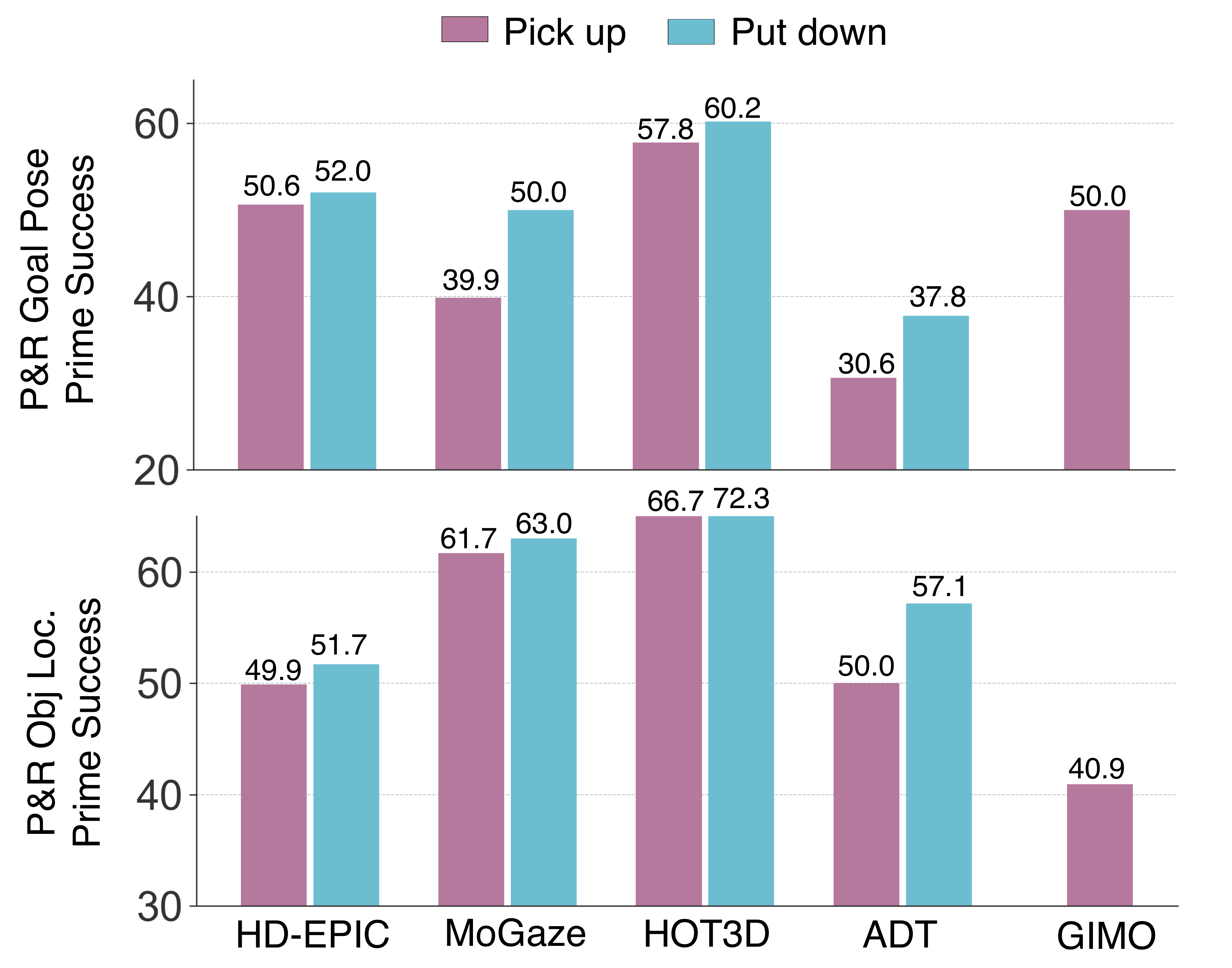}
        \captionof{figure}{\pr\ performance for pick v/s put.}
        \label{fig:pick_and_put_plot}
\end{minipage}
\end{table*}

\subsection{Baselines}
\label{sec:comparison}

To assess the ability of generated motion to replicate humans' \pr\ behaviour, we benchmark six methods on our curated datasets: one naive baseline and five previous works (one from the text-to-motion generation and four from location-conditioned human motion generation):
\begin{itemize}[leftmargin=*,noitemsep, nolistsep,label=\textbullet]
 \item \textbf{Static} is a naive baseline that uses the average full-body pose of training data and keeps it static. It showcases the difficulty of the dataset.
 \item \textbf{MDM}~\cite{tevet2023mdm}. We evaluate the checkpoint trained on HumanML3D~\cite{Guo_2022_CVPR} vs our pre-training on Nymeria~\cite{ma2024nymeria}. We also fine-tune this model, pre-trained on Nymeria, on our training split using only text conditioning.
 \item \textbf{GMD}~\cite{karunratanakul2023gmd}, a guided motion diffusion trained on HumanML3D, is a two-stage motion generation method. The first stage generates a root trajectory that guides full-body motion generation in the second stage.
 \item \textbf{DNO}~\cite{karunratanakul2024optimizing} is a framework that treats the pre-trained text-to-motion diffusion model as motion priors and optimises the starting latent noise through backpropagation of task-specific gradients without training for each new task.
 \item \textbf{DartControl}~\cite{Zhao:DartControl:2025} autoregressively predicts motion primitives from text and past motion. We adapt its original in-betweening scheme for goal pose and location conditioning, optimising latent noise via DNO for 100 steps.

 \item \textbf{WANDR}~\cite{diomataris2024wandr} is an autoregressive c-VAE trained on AMASS and CIRCLE for frame-by-frame motion generation. It utilizes \textit{intention features} to encode goal location and remaining time.
\end{itemize}

\subsection{Results}
\label{sec:quantitative}

In \cref{table:baseline_comparison}, we report results of one model trained on all training sequences from different source datasets. We compare our proposed \pr\ diffusion model against baselines, reporting results separately on each test set.
In Supplementary, we report analogous results where we train on each dataset independently.

The naive static baseline performs poorly on all the metrics for all datasets. Its poor prime success (no priming) and reach success ($12\%$ on average) highlight the difficulty of the task on our curated datasets.

Text-conditioned baselines rely solely on the knowledge of the pick/put action to generate motion and have no information about the target location to prime or reach. 
We report the performance of three variants of MDM~\cite{tevet2023mdm}.
Text-conditioned baselines perform poorly on all the metrics as they lack sufficient guidance for goal location. Even a fine-tuned MDM model on the \pr\ sequences has poor performance on prime and reach success on most of the datasets, showing that text is not a sufficient condition for prime and reach.

\noindent \textbf{Conditioning with Goal Pose}.
We compare our method with three recent controllable motion generation approaches:
GMD~\cite{karunratanakul2023gmd}, DNO~\cite{karunratanakul2024optimizing}, and DartControl~\cite{Zhao:DartControl:2025}, each evaluated both in a zero-shot setting and after fine-tuning on our \pr\ sequences.
In contrast, our goal-pose conditioned \pr\ model consistently achieves the highest prime and reach success across all five datasets, with gains of up to $+18.2\%$ in prime success on GIMO and $+25.3\%$ in reach success on MoGaze over the best goal-pose baseline.

\begin{figure}[t]
    \centering
    \includegraphics[width=\linewidth]{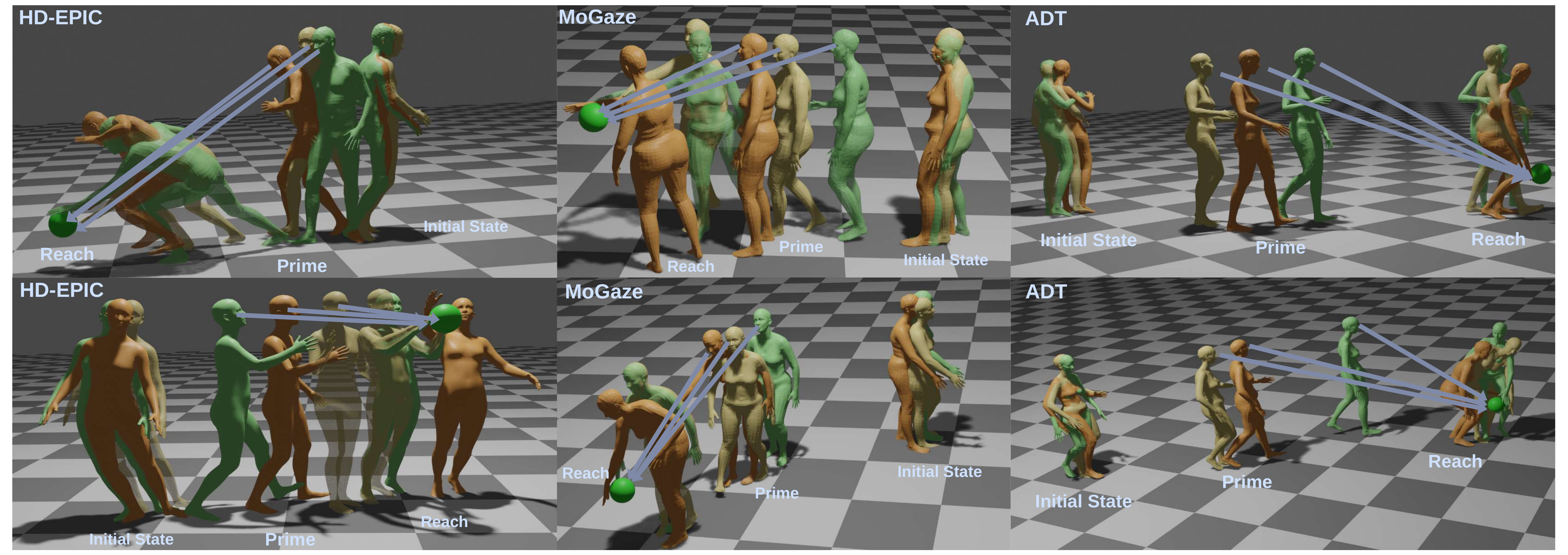}
    \caption{Qualitative results on 3 datasets: Ground truth sequence in \textcolor[rgb]{0.329, 0.8, 0.26}{light green}, goal-pose conditioned prediction in \textcolor[rgb]{0.8, 0.653, 0.243}{translucent yellow}, and target location conditioned generation in \textcolor[rgb]{0.834, 0.353, 0.091}{brown}. We show the pose at the initial, prime, and reach timesteps. Prime direction for both ground truth and predictions is shown using \textcolor[HTML]{7F8AA9}{arrows}, and target object location is shown in \textcolor[rgb]{0.052, 0.8, 0.042}{sphere}.}
   \vspace*{-12pt}\label{fig:qualitative}
\end{figure}

\noindent \textbf{Conditioning with Object Location}.
For object location conditioning, we compare with WANDR~\cite{diomataris2024wandr}, DNO~\cite{karunratanakul2024optimizing}, and DartControl~\cite{Zhao:DartControl:2025}, where each model was optimised or trained so that the right wrist of the final frame reaches the target location.
Similar to goal pose conditioned baselines, we evaluate these methods in a zero-shot setting and after fine-tuning on our dataset.
By construction, latent-noise optimisation methods, used in DNO, DartControl, and \pr\, achieve near-perfect reach success across most datasets, since the optimisation objective focuses on the final wrist-target distance.
In most cases, \pr\ outperforms zero-shot and fine-tuned baselines on all metrics. 
Without guidance on the goal pose, the location error increases for all models.
\pr\ achieves the lowest location error in 4 out of the 5 datasets (2nd best on HOT3D).
\pr\ achieves the best prime success on all datasets.

\noindent \textbf{Impact of fine-tuning on \pr\ sequences}.
For both goal-pose and object-location conditioned generation, all baselines generally benefit from fine-tuning on our \pr\ sequences, resulting in higher prime and reach success on most datasets. In particular, GMD achieves average improvements of $10.2\%$ and $8.1\%$ in prime success and reach success, respectively. 
This highlights the usefulness of our curated \pr\ sequences for natural human motion generation.

\noindent \textbf{Why \pr\ outperforms optimisation-based baselines}.
Unlike prior methods that mainly use the goal pose or object location as an optimisation target on the final frame, \pr\ is trained to generate full trajectories conditioned on both the initial state and the goal, encouraging it to learn natural priming strategies (\ie, how to move head, torso, and body to first attend to the future interaction location) rather than merely steering the endpoint. As a result, \pr\ achieves substantially higher prime success while maintaining strong reach performance.

\noindent \textbf{Results for pick and put}.
We analyse the performance of the \pr\ model separated by the action (\ie, pick up or put down) in \cref{fig:pick_and_put_plot}.
Overall, pick-up motions are relatively more challenging than put-down actions, especially for priming ability on MoGaze when conditioned on the goal pose.

\noindent \textbf{Qualitative Results}.
We demonstrate qualitative examples of our \pr\ generated motions in \cref{fig:qualitative}. Generated \pr\ motions appear natural. Starting with an initial pose and velocity, our generated motion first primes the target object (see arrow) and then reaches it with one of the hands. 
Evidently, using the goal location matches better the ground-truth. However, using the object location condition solely successfully synthesises reach but positions the body in a different location at the goal.
We demonstrate this ability on various target locations, including challenging ones, where the location can be located low or high. 

\begin{table}[t!]
\centering
\caption{\textbf{Impact of condition}: We show how each of our modified conditions and optimisation impacts \pr\ model's performance (last row).}
\vspace*{-6pt}
\label{tab:condition_ablation}
\resizebox{\linewidth}{!}{
\begin{tabular}{ccccc|cccc|cccc}
\toprule
&&&&& \multicolumn{4}{c|}{HD-EPIC} & \multicolumn{4}{c}{MoGaze}\\
\midrule
\begin{tabular}[c]{@{}c@{}}Object \\ Loc.\end{tabular}&\begin{tabular}[c]{@{}c@{}}Initial \\ Pose\end{tabular} & \begin{tabular}[c]{@{}c@{}}Initial \\ Vel.\end{tabular}  & Text & $\mathcal{L}_{\text{opt}}$ &\begin{tabular}[c]{@{}c@{}}Prime \\ Success\end{tabular} $\uparrow$ & \begin{tabular}[c]{@{}c@{}}Reach \\ Success\end{tabular} $\uparrow$  & \begin{tabular}[c]{@{}c@{}}Loc \\ Err\end{tabular} $\downarrow$  & MPJPE $\downarrow$&\begin{tabular}[c]{@{}c@{}}Prime \\ Success\end{tabular} $\uparrow$ & \begin{tabular}[c]{@{}c@{}}Reach \\ Success\end{tabular} $\uparrow$ & \begin{tabular}[c]{@{}c@{}}Loc \\ Err\end{tabular} $\downarrow$  & MPJPE $\downarrow$\\
\midrule
\xmark & \checkmark& \checkmark& \checkmark& \xmark&28.50 & 30.62& 65.50& 0.45 & 13.75 & 10.94 &  73.21 & 0.79 \\
\checkmark & \xmark & \xmark  & \checkmark &\xmark&39.61 &79.64 & 31.30 & 0.38 &50.67 & 81.15 &  49.57 & 0.69\\
\checkmark &\xmark & \checkmark & \checkmark&\xmark&42.25 &81.15 & 28.41&0.40& 52.80 & 87.57 &  42.67 & 0.71 \\
\checkmark &\checkmark& \xmark &\checkmark & \xmark &46.90 &84.31 & 29.00& 0.38& 56.72 & 91.57  & 41.59 & 0.65  \\
\checkmark&\checkmark&\checkmark & \xmark & \xmark & 48.78 & 88.46&27.55 &0.30&57.34 & 92.52  & 37.79 & 0.60\\
\checkmark&\checkmark&\checkmark & \checkmark & \xmark &49.60&88.68&27.30&0.30&59.67 & 92.64 & 37.70 & 0.59 \\
\bottomrule 
\rowcolor{LightRed}\checkmark&\checkmark&\checkmark & \checkmark & \checkmark &\textbf{51.00}&\textbf{100.00}&\textbf{25.26}&\textbf{0.27}&\textbf{62.37} & \textbf{100.00} & \textbf{36.43} & \textbf{0.56} \\
\bottomrule 
\end{tabular}}
\end{table}

\begin{table}[t!]
\centering
\caption{\textbf{Impact of pre-training}. To validate our pre-training on Nymeria, we show the \pr\ model's performance without pre-training and pre-trained on HumanML3D.}
\label{tab:pretraining_impact}
\vspace*{-8pt}
\resizebox{\linewidth}{!}{
\begin{tabular}{lcccc|cccc}
\toprule
& \multicolumn{4}{c|}{HD-EPIC} & \multicolumn{4}{c}{MoGaze}\\
\midrule
Pre-train & \begin{tabular}[c]{@{}c@{}}Prime \\ Success\end{tabular} $\uparrow$ & \begin{tabular}[c]{@{}c@{}}Reach \\ Success\end{tabular} $\uparrow$   & Loc Err $\downarrow$ & MPJPE $\downarrow$ & \begin{tabular}[c]{@{}c@{}}Prime \\ Success\end{tabular} $\uparrow$ & \begin{tabular}[c]{@{}c@{}}Reach \\ Success\end{tabular} $\uparrow$   & Loc Err $\downarrow$ & MPJPE $\downarrow$ \\
\midrule
No pre-train&41.20 &98.86&42.18&0.36& 49.50 & 99.50 & 48.11 & 0.68 \\
HumanML3D~\cite{Guo_2022_CVPR} &49.90 &\textbf{100.00}&32.76&0.30& 57.45 & \textbf{100.00} & 40.85 & 0.59 \\
\midrule
\rowcolor{LightRed}Nymeria (Ours) & \textbf{51.00}&\textbf{100.00}&\textbf{25.26}&\textbf{0.27}&\textbf{62.37} & \textbf{100.00} & \textbf{36.43} & \textbf{0.56} \\
\bottomrule 
\end{tabular}}
\vspace*{-12pt}
\end{table}

\subsection{Ablation and Analysis}
\label{sec:ablation}
We ablate the proposed \pr\ model on our two largest datasets: HD-EPIC and MoGaze. These cover both estimated motion (using EgoAllo for HD-EPIC) and MoCap data (in MoGaze).

\noindent \textbf{Condition Ablation}.
As explained in \cref{sec:prime_and_reach_method}, the proposed \pr\ method uses text, initial state (pose and velocity), and target location as conditions. We ablate the impact of each of these conditions in \cref{tab:condition_ablation}. Using the object location condition gives a significant boost in all metrics, with maximum gains of $81.7\%$ for reach success and $35.5\%$ in location error on MoGaze. This showcases the difficulty of the task, and that it is not plausible to synthesise \pr\ motions without knowledge of the target. Using the initial state of the body as a condition helps to improve the priming ability of the generated motion, leading to a gain of $10.0\%$ in prime success on HD-EPIC. We find that having both initial pose and initial velocity as initial state conditions is important, especially for the prime success, with drops of at least $2.7\%$ when either is removed on HD-EPIC. Finally, the ablations show that having action knowledge via text (\ie, drop or pick) also improves \pr\ motion generation on most metrics.

\noindent \textbf{Impact of latent noise optimisation $\mathcal{L}_{\text{opt}}$}.
As noted in prior work, adopting this optimisation notably improves performance across metrics (except for prime success in MoGaze, which drops marginally).

\noindent \textbf{Impact of pre-training}.
We find that the Nymeria pre-trained model gives a better initialisation (\cref{tab:pretraining_impact}) thanks to its diverse, natural human interactions with objects, including intentional interaction motions.

\section{Conclusion and Future Work}
Humans naturally spot or prime an object before reaching it. Previous motion synthesis benchmarks or methods have failed to explore the role of priming for object reaching. 
To that end, we curate Prime and Reach (\pr) sequences from five datasets using gaze information and object locations. We propose a \pr\ motion diffusion model that generates full-body motion using goal pose or target location as a condition, along with initial state and text conditioning. 
Results demonstrate improved generation compared to prior baselines.

\noindent \textbf{Limitation} Similar to other works~\cite{diomataris2024wandr,Zhao:DartControl:2025,karunratanakul2024optimizing}, we do not model hand pose (only body up to wrist). Generating hand motion is an interesting future direction due to its relevance to grasping objects upon reach. 

\noindent\textbf{Acknowledgements:} This work uses publicly available datasets and annotations to curate \pr\ sequences. Research at the University of Bristol is supported by EPSRC UMPIRE
(EP/T004991/1).
M Hatano is supported by JST BOOST, Japan
Grant Number JPMJBS2409, and Amano Institute of
Technology.
S Sinha and J Chalk are supported by EPSRC DTP studentships.

At Keio University, we used ABCI 3.0 provided by AIST and AIST Solutions.

At the University of Bristol, we acknowledge the use of resources provided by the Isambard-AI National AI Research Resource (AIRR), funded by the UK Government’s Department for Science, Innovation and Technology (DSIT) via UK Research and Innovation; and the Science and Technology Facilities Council [ST/AIRR/I-A-I/1023].
In particular, we acknowledge the usage of GPU Node hours granted as part of the AIRR Gateway project ``HOI Foundational Model from Egocentric Data'' (Dec 2025 - Mar 2026), Sovereign AI Unit call project ``Gen Model in Ego‑sensed World'' (Aug-Nov 2025) as well as the usage of GPU Node hours granted by AIRR Early Access Project ANON-BYYG-VXU6-M (March-May 2025).

%
%
\bibliographystyle{splncs04}
\bibliography{main}
\renewcommand{\thefigure}{S\arabic{figure}}
\renewcommand{\thetable}{S\arabic{table}}
\setcounter{figure}{0}
\setcounter{table}{0}
\newcommand{\tighttoc}{
    \begingroup
    \renewcommand{\clearpage}{}
    \renewcommand{\newpage}{}
    \tableofcontents
    \endgroup
}

\clearpage

\begin{center}
\Large{\textbf{Prime and Reach: Synthesising Body Motion for Gaze-Primed Object Reach}}\\
\textmd{-- Supplementary Materials --}
\end{center}

\appendix

\addtocontents{toc}{\protect\setcounter{tocdepth}{2}}
\hypersetup{linkcolor=black}
\tighttoc
\hypersetup{linkcolor=blue}

\section{Qualitative Videos}
\label{sec:qualitative_video}
We include the qualitative videos on our website \url{https://masashi-hatano.github.io/prime-and-reach/} showcasing predicted motion sequences from our \pr\ model over different datasets. For each of the sequences, we provide the goal location in \textcolor[rgb]{0.052, 0.8, 0.042}{green sphere}, our goal-pose conditioned prediction in \textcolor[rgb]{0.8, 0.653, 0.243}{yellow}, the goal-location conditioned synthesis in \textcolor[rgb]{0.834, 0.353, 0.091}{brown} and corresponding ground truth motion in \textcolor[rgb]{0.329, 0.8, 0.265}{green}.

\section{Further Details on \pr\ sequence curation}
\subsection{Slab Test Method for Priming}

The Slab Test Method expects the knowledge of the target location $o_{\text{3D}}$, which is an axis-aligned 
3D bounding box, 
defined by its minimum ($\mathbf{b}_\text{min}$) and maximum ($\mathbf{b}_\text{max}$) corners, or as 3D coordinates of the object center. 

\begin{figure*}[t]
    \centering
    \includegraphics[width=\linewidth]{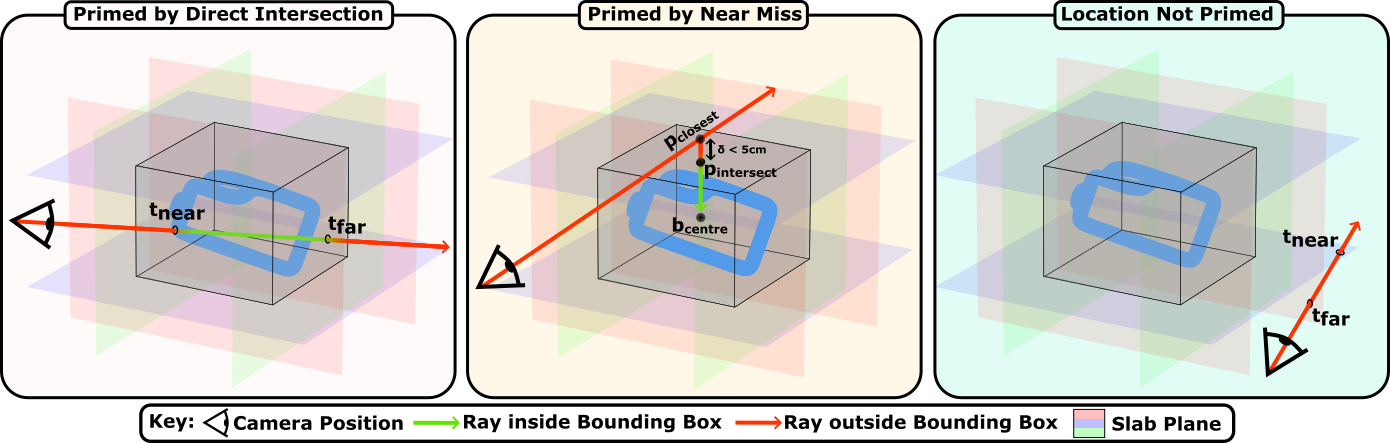}
    \caption[The Slab Test Method for Priming Objects]{Visualisation of the Slab Test Method~\cite{Majercik2018Voxel} for registering primed object interactions. (Left) Primed by Valid Intersection. (Middle) Primed by Near Miss. (Right) No Priming.}
    \label{fig:slab_test_method}
\end{figure*}

The Slab Test Method treats the box as the overlapping volume of three infinite slabs (one for each axis), each bounded by a pair of parallel planes. A visualisation of the intersection checks is shown in \cref{fig:slab_test_method}. The algorithm calculates two key parametric distances along the gaze ray. The first, $t_{\text{near}}$, represents the distance to the last slab plane that the ray enters. It is the furthest entry point, marking the moment the ray is inside all three slabs and thus inside the box. The second, $t_{\text{far}}$, is the distance to the first slab plane that the ray exits. It is the nearest exit point, marking the moment the ray leaves the box volume. A valid intersection occurs if the ray enters the box before it exits, as defined by the condition in Equation~\ref{eq:aabb_intersect}, 

\begin{gather}
    t_\text{near} = \max_{i \in {x,y,z}} \min\left(\frac{\mathbf{b}^{(i)}_\text{min} - \mathbf{o}^{(i)}_\text{cam}}{\hat{\mathbf{d}}^{(i)}_\text{gaze}}, \frac{\mathbf{b}^{(i)}_\text{max} - \mathbf{o}^{(i)}_\text{cam}}{\hat{\mathbf{d}}^{(i)}_\text{gaze}}\right) \nonumber \\
t_\text{far} = \min_{i \in {x,y,z}} \max\left(\frac{\mathbf{b}^{(i)}_\text{min} - \mathbf{o}^{(i)}_\text{cam}}{\hat{\mathbf{d}}^{(i)}_\text{gaze}}, \frac{\mathbf{b}^{(i)}_\text{max} - \mathbf{o}^{(i)}_\text{cam}}{\hat{\mathbf{d}}^{(i)}_\text{gaze}}\right) \nonumber \\
    \text{Intersection if } t_\text{near} < t_\text{far} \text{ and } t_\text{far} \ge 0,
\label{eq:aabb_intersect}
\end{gather}
where $\mathbf{o}_\text{cam}$ and $\hat{\mathbf{d}}_\text{gaze}$ denote the location of the camera and direction of gaze originating from the camera, respectively.

To account for near misses where gaze is directed towards an object but does not intersect its bounding box, we employ a proximity check. First, for a given gaze ray originating at $\mathbf{o}_\text{cam}$ with direction $\hat{\mathbf{d}}_\text{gaze}$, we find the point on the ray, $\mathbf{p}_\text{closest}$, that has the minimum distance to the centre of the object's 3D bounding box, $\mathbf{b}_\text{centre}$. This point is found by projecting the vector from the camera to the box centre onto the gaze ray, as shown in Equation~\ref{eq:closest_point}.
\begin{gather}
t_\text{closest} = (\mathbf{b}_\text{centre} - \mathbf{o}_\text{cam}) \cdot \hat{\mathbf{d}}_\text{gaze} \nonumber \\
\mathbf{p}_\text{closest} = \mathbf{o}_\text{cam} + t_\text{closest}\cdot \hat{\mathbf{d}}_\text{gaze}
\label{eq:closest_point}
\end{gather}
From this closest point, we cast a new ray directly towards the bounding box centre, $\hat{\mathbf{d}}_\text{centre}$, and use the slab test method to identify where this new ray intersects the box. Specifically, we swap $\mathbf{o}_\text{cam}$ for $\mathbf{b}_\text{centre}$ and $\hat{\mathbf{d}}_\text{gaze}$ for $\hat{\mathbf{d}}_\text{centre}$ in Equation~\ref{eq:aabb_intersect}, yielding a point:
\begin{equation}
\mathbf{p}_\text{intersect} = \mathbf{b}_\text{centre} + t_\text{near}\cdot\hat{\mathbf{d}}_\text{centre}
\end{equation}
A location is considered primed by a near miss if the Euclidean distance, $\delta$, between $\mathbf{p}_\text{closest}$ and $\mathbf{p}_\text{intersect}$ is below a threshold $\tau$ of 5 cm. This threshold was determined empirically: we found that smaller values risked undercounting valid gaze interactions due to minor inaccuracies in gaze or object bounding boxes, while larger values began to accept ambiguous cases. Formally, priming by near miss occurs when:

\begin{gather}
\delta = ||\mathbf{p}_\text{intersect} - \mathbf{p}_\text{closest}|| \nonumber \\
\text{Near miss if } \delta \le \tau \text{ and } t_\text{closest} \ge 0
\label{eq:near_miss}
\end{gather}

The second condition in Equation~\ref{eq:near_miss} ensures the closest point lies in front of the camera, confirming the user is looking towards the object. 

We exclude interactions involving only minimal movement ($<$ 20 cm) between the initial pose and goal, as they do not represent meaningful interactions. This filtering process refines the dataset and ensures the quality of \pr\ sequences so that primed object interactions are not trivial.

\subsection{Estimating Full Body Pose for \pr\ Sequences}
\label{sec:generating_sequences}
Building upon the priming data collected previously, we require full-body pose sequences of primed object interactions.
Our generation pipeline uses EgoAllo~\cite{yi2025egoallo}, a method that estimates expressive, full-body human motion from egocentric video and SLAM-based camera poses. The model first converts head pose trajectories into a spatially and temporally invariant representation that encodes relative motion with respect to the ground plane. This representation is used to condition a diffusion-based prior that samples local SMPL-H~\cite{MANO:SIGGRAPHASIA:2017} parameters: pose, representing per-joint rotations over time for the full body including hands; shape, encoding time-invariant body proportions such as height and limb length; and contact predictions, indicating per-joint contact with the environment to improve realism. The model is trained on human motion sequences from AMASS~\cite{mahmood2019amass}, augmented with synthetic egocentric head pose trajectories.

For each interaction, we provide the model with a sequence of video frames and their corresponding camera poses to generate an initial sequence of full-body motions.
To enhance the fidelity of hand-object interactions, we estimate the 3D wrist and palm poses from Aria MPS models and provide these to the EgoAllo model to align the generated hands with the wrist and hand locations.
We found that without this alignment step, the fidelity of the hands in the generated sequence is often diminished. 
Incorporating these poses yields a more accurate representation of hand positions and their orientations in our final motion sequences.

A key design choice in our generation process is the temporal window of the sequences. Specifically, we initiate the generation 2 seconds prior to the moment the object is primed and conclude following the interaction. This decision was made to ensure that our sequences capture any sufficient head motions or other preparatory body movements that precede the explicit eye-gaze priming. By including this anticipatory phase, the resulting sequences provide a more complete and naturalistic depiction of a primed interaction.

\subsection{Comparing EgoAllo to Mocap.}
To verify the suitability of EgoAllo outputs as an approximation to the standard 3D body pose annotations, typically acquired using Mocap, we compare the results of \pr\ when using EgoAllo in place of the Mocap data on the ADT~\cite{pan2023aria} dataset. 
We chose this dataset as it uses the Aria glasses, making it suitable for EgoAllo body pose estimations.
One thing to note is that hand (palm and wrist) tracking data is not available from the Aria glasses in this dataset, which are known to improve the EgoAllo estimations.

The results are shown in~\cref{table:egoallo_vs_mocap}.
We provide results when training using Mocap and EgoAllo as well as when evaluating on the test set poses from Mocap (left) and EgoAllo (right).
Overall, \pr\ trained on EgoAllo body pose estimates performs slightly worse than the corresponding ground truth.
When conditioning with object location, EgoAllo performs similarly to Mocap (\eg 97.64\% vs.\ 100\% Reach Success). While there is a performance drop in Reach Success under the goal pose condition when evaluating on the Mocap test set (51.76\% vs.\ 70.58\%), the broader motion-based metrics (\eg Goal MPJPE, MPJPE, Loc Err and Foot Skating) still remain comparable.
For example, MPJPE drops by $< 0.1m$ in every case when comparing models trained on EgoAllo to those trained on Mocap.
This indicates that the overall motion quality is acceptable.

\begin{table}[!t]
\caption{\textbf{Comparison of EgoAllo outputs and ground truth.} We compare our \pr\ method on ADT~\cite{pan2023aria} when trained and evaluated on both ground-truth Mocap data and EgoAllo.}
\label{table:egoallo_vs_mocap}
\resizebox{\textwidth}{!}{
    \begin{tabular}{cccccccc|cccccc}
        \toprule
        & & \multicolumn{6}{c|}{ADT (Mocap)}& \multicolumn{6}{c}{ADT (EgoAllo)}\\
        \midrule
        Condition & \begin{tabular}[c]{@{}c@{}}Training \\ Data\end{tabular} & \begin{tabular}[c]{@{}c@{}}Prime \\ Success\end{tabular} $\uparrow$ & \begin{tabular}[c]{@{}c@{}}Reach \\ Success\end{tabular} $\uparrow$ & \begin{tabular}[c]{@{}c@{}}Goal \\ MPJPE\end{tabular} $\downarrow$  & Loc Err $\downarrow$ &  MPJPE $\downarrow$ & \begin{tabular}[c]{@{}c@{}}Foot \\ Skating\end{tabular} $\downarrow$ & \begin{tabular}[c]{@{}c@{}}Prime \\ Success\end{tabular} $\uparrow$&\begin{tabular}[c]{@{}c@{}}Reach \\ Success\end{tabular} $\uparrow$ & \begin{tabular}[c]{@{}c@{}}Goal \\ MPJPE\end{tabular} $\downarrow$    & Loc Err $\downarrow$  & MPJPE $\downarrow$ & \begin{tabular}[c]{@{}c@{}}Foot \\ Skating\end{tabular} $\downarrow$ \\
        \midrule
        \multirow{2}{*}{\begin{tabular}[l]{@{}l@{}}+ Initial State \\ \& Goal Pose\end{tabular}}
        & EgoAllo &
        29.41 & 51.76 & 0.17 & 5.88 & 0.38 & 0.17 & 28.05 & 81.17 & 0.18 & 4.87 & 0.42 & \textbf{0.18}  \\
        & Mocap &
        \textbf{37.64} & \textbf{70.58} & \textbf{0.08} & \textbf{1.17} & \textbf{0.33} & \textbf{0.11} & \textbf{31.70} & \textbf{92.68} & \textbf{0.13} & \textbf{1.22} & \textbf{0.38} & 0.21   \\
        \midrule
        \multirow{2}{*}{\begin{tabular}[l]{@{}l@{}}+ Initial State \\ \& Object Loc.\end{tabular}}
        & EgoAllo &
        \textbf{45.88} & 97.64 & \textbf{0.55} & 52.94 & 0.54 & 0.18 & 36.58 & 96.34 & \textbf{0.53} & \textbf{45.12} & 0.54 & \textbf{0.18}  \\
        & Mocap &
        43.53 & \textbf{100.00} & \textbf{0.55} & \textbf{52.54} & \textbf{0.48} & \textbf{0.11} & \textbf{39.02} & \textbf{100.00} & 0.54 & \textbf{45.12} & \textbf{0.51} & 0.20  \\
        \bottomrule
    \end{tabular}
}
\end{table}

\subsection{More Statistics of \pr\ sequences}
\begin{figure*}[!t]
    \centering
    \includegraphics[width=\linewidth]{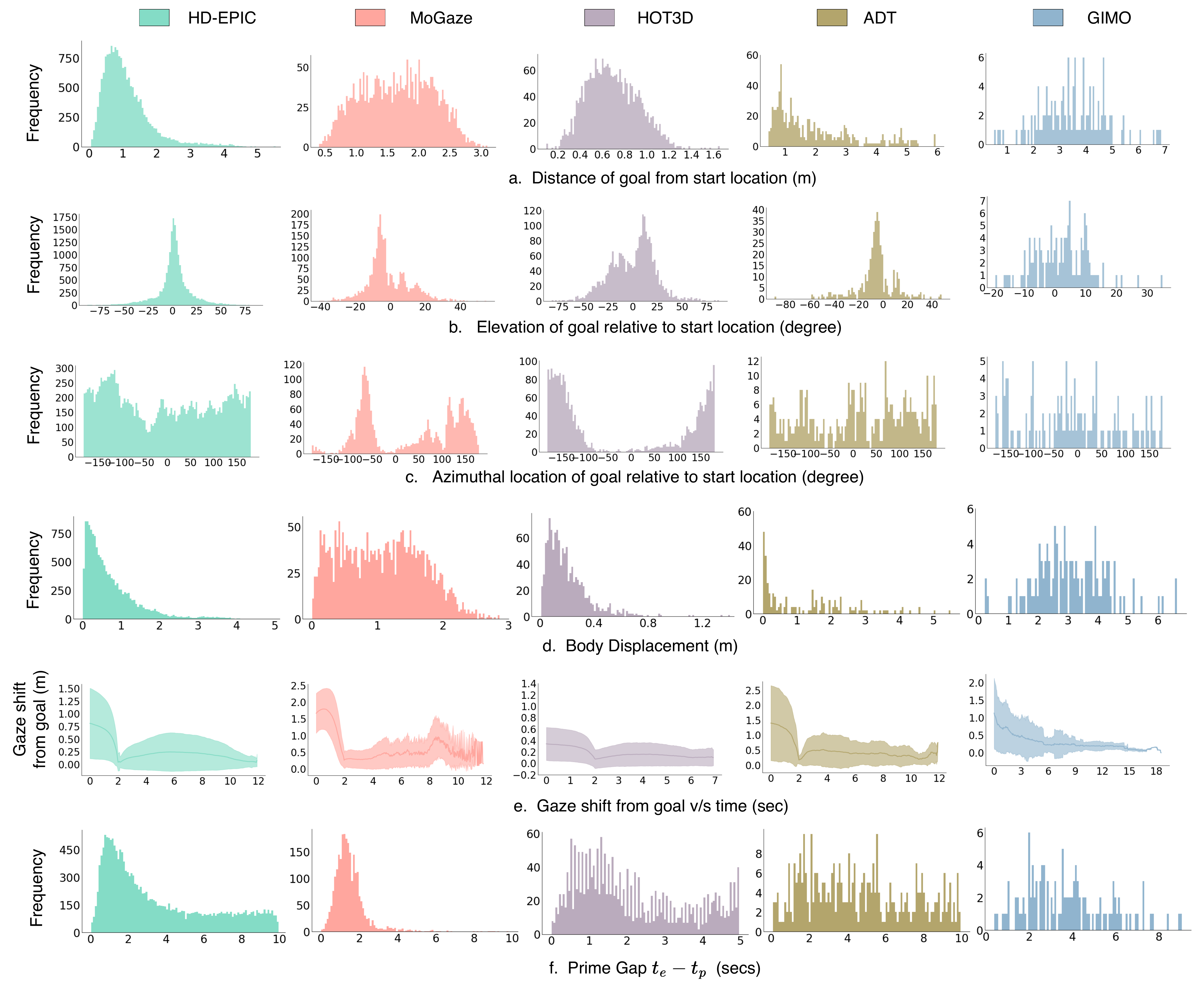}
    \caption{Statistics of the curated \pr\ sequences. In a, b, and c, we highlight the high diversity of our goal locations by plotting the histogram of goal distance, elevation, and azimuthal location relative to the start location. In d, we plot the histogram of body movement in our sequences.  In e, we plot how the gaze shifts from the goal during the prime and reach motion, where the shift is minimum when the gaze is on the object. Finally in f, we plot the histogram for time interval between priming and reaching.}
    \label{fig:more_stats}
\end{figure*}

We show more detailed statistics on each of our curated datasets in \cref{fig:more_stats}. Concretely, the histograms of body movement, hand movement, and prime gap are shown. Body and hand movement measure the maximum displacement of the body or hands within a \pr\ motion sequence. Prime gap is the duration between the prime time $t_{p}$ and the pick/put event time $t_{e}$. 

\subsection{Train-Test Splits}
For each dataset, we split the source videos into 70\% train -30 \% test sets. The \pr\ sequences curated from these videos were automatically distributed to the corresponding subset. HD-EPIC \cite{perrett2025HD-EPIC} has 156 long videos. We selected 70\% (109 videos) for training and the remaining for testing. The curated sequences from the 109 videos were used as train \pr\ sequences. We perform a similar procedure for MoGaze~\cite{kratzer2020mogaze}, HOT3D~\cite{banerjee2025hot3d}, and ADT~\cite{pan2023aria}. 
Zheng \etal \cite{zheng2022gimo} proposed a train-test split for GIMO sequences. We use the same split for our curated \pr\ sequences.
Exact train/test split sizes are given in \cref{tab:train_test_splits}.

\begin{table}[t!]
\centering
\caption{\textbf{Train/Test splits}. We provide the train-test splits for our curated \pr\ sequences.}
\vspace*{-8pt}
\tiny
\resizebox{0.5\linewidth}{!}{
\begin{tabular}{lccccc}
\toprule
 & HD-EPIC & MoGaze & HOT3D & ADT & GIMO\\
\midrule
Train & 12642 & 1947 &1672& 326 & 108 \\
Test & 5492& 690 & 744& 85 & 22\\
\midrule
Total & 18134 & 2637 & 2416& 411 & 130\\
\bottomrule 
\end{tabular}
}
\label{tab:train_test_splits}
\end{table}

\section{Per-Dataset Models}
In the main paper, we train a single model on the combined training sequences of all datasets, and present our main results in Tab. 2.
For completion, and to further validate the effectiveness of our proposed method, we evaluate the performance of the models trained and tested independently (\ie separately) on each of the five datasets - training one model per dataset.

As shown in \cref{table:baseline_comparison_per_Dataset}, the performance trends align closely with the results reported in the main manuscript. 
Even when trained on domain-specific data, the naive static and text-conditioned baselines perform poorly across all metrics. 
This confirms that the difficulty of the \pr\ task, specifically the need for spatial guidance and anticipatory motion, cannot be overcome by restricting training to a single dataset if the model architecture lacks sufficient conditioning cues.
For conditioning with goal pose, our \pr\ model performs the highest prime success in all datasets and reach success in 4 out of 5 datasets. In particular, \pr\ outperforms the strongest baselines by $15.1\%$ and $19.9\%$ in prime success and reach success on the MoGaze dataset, respectively.
A similar trend can be seen in the results with object location conditioning, where the proposed method achieves the highest prime success in most datasets while keeping reach success nearly perfect.
Furthermore, our model consistently yields the lowest location error compared to all baselines.
Taken together, these results demonstrate that the proposed method is robust across diverse domains and independent of the size of the training data.

\begin{table}[!htp]
\caption{\textbf{Per-Dataset Training - Comparison of motion generation baselines.} Here, we train an independent model per dataset (HD-EPIC, MoGaze, HOT3D, ADT, and GIMO), and report results for each. The baselines are grouped by the type of conditioning used for generation. $\dagger$ denotes the zero-shot inference. For MDM, we evaluate two pre-trained models: (1) trained on HumanML3D~\cite{Guo_2022_CVPR}, and (2) trained on Nymeria.\cite{ma2024nymeria}, denoted as $\ddagger$ and $^*$, respectively. Entries without a marker correspond to models fine-tuned on our per-dataset \pr\ sequences.}
\label{table:baseline_comparison_per_Dataset}
\begin{subtable}[t]{\linewidth}
\resizebox{\textwidth}{!}{
\begin{tabular}{cccccccc|cccccc}
\toprule
& & \multicolumn{6}{c|}{HD-EPIC}& \multicolumn{6}{c}{MoGaze}\\
\midrule
Condition& Method& \begin{tabular}[c]{@{}c@{}}Prime \\ Success\end{tabular} $\uparrow$ & \begin{tabular}[c]{@{}c@{}}Reach \\ Success\end{tabular} $\uparrow$ & \begin{tabular}[c]{@{}c@{}}Goal \\ MPJPE\end{tabular} $\downarrow$  & Loc Err $\downarrow$ &  MPJPE $\downarrow$ & \begin{tabular}[c]{@{}c@{}}Foot \\ Skating\end{tabular} $\downarrow$ & \begin{tabular}[c]{@{}c@{}}Prime \\ Success\end{tabular} $\uparrow$&\begin{tabular}[c]{@{}c@{}}Reach \\ Success\end{tabular} $\uparrow$ & \begin{tabular}[c]{@{}c@{}}Goal \\ MPJPE\end{tabular} $\downarrow$    & Loc Err $\downarrow$  & MPJPE $\downarrow$ & \begin{tabular}[c]{@{}c@{}}Foot \\ Skating\end{tabular} $\downarrow$ \\
\midrule
No condition & Static & 0 & 23.16 & 0.70 & 50.91 & 0.45 & \multicolumn{1}{c|}{--} & 0 & 2.56 & 1.06 & 75.99 & 0.62 & -- \\
\midrule
\multirow{3}{*}{Text} & MDM $\dagger \ddagger$ &9.53& 13.94 &1.14&81.75 &0.96 & \multicolumn{1}{c|}{0.16} & 2.85 & 2.20 & 2.03 & 94.11 & 1.45 & 0.39 \\
& MDM $\dagger ^*$  & 9.52& 13.47 &0.85&62.13 &0.59 &\multicolumn{1}{c|}{0.06} & 3.11 & 1.85 & 1.19 & 79.76 & 0.73 & 0.06 \\
& MDM &
14.30&  19.85  & 0.82 & 52.26& 0.51 & \multicolumn{1}{c|}{0.06} & 6.81 & 4.50 & 1.19 & 82.62 & 0.74 & 0.04 \\
\midrule
\multirow{7}{*}{\begin{tabular}[l]{@{}l@{}}+ Initial State \\ \& Goal Pose\end{tabular}}
& GMD~$\dagger$\cite{karunratanakul2023gmd} & 33.27& 30.77 & 0.26 & 2.00 & 0.32& \multicolumn{1}{c|}{0.10} & 8.11 & 10.86 & 0.35 & 2.23 & 0.54 & 0.11 \\
& GMD\cite{karunratanakul2023gmd} & 
44.40& 56.74  & 0.21 & 2.00 & 0.30& \multicolumn{1}{c|}{0.06} & 25.38 & 34.66 & 0.28 & 1.00 & 0.53 & 0.21 \\
& DNO~$\dagger$\cite{karunratanakul2024optimizing} & 43.42 & 87.44 & \textbf{0.07} & 2.60 & 0.30 & \multicolumn{1}{c|}{0.04} & 17.39 & 69.71 & 0.09 & 6.81 & 0.64 & \textbf{0.07} \\
& DNO~\cite{karunratanakul2024optimizing} &
48.99 & 89.16 &0.09  & \textbf{0.49} & 0.25 & \multicolumn{1}{c|}{0.06} & 31.15 & 78.68 & 0.12 & 0.48 & 0.55 & 0.18 \\
& DartControl~$\dagger$~\cite{Zhao:DartControl:2025} & 30.06 & 81.23 & 0.14 & 2.20 & 0.43 & \multicolumn{1}{c|}{0.15} & 38.26 & 71.16 & 0.11 & 0.14 & 0.52 & 0.35 \\
& DartControl~\cite{Zhao:DartControl:2025} & 34.65 & 88.46 & 0.11 & 1.78 & 0.37 & \multicolumn{1}{c|}{0.09} & 34.78 & 65.94 & 0.12 & 0.14 & 0.58 & 0.51\\

\rowcolor{LightRed}
\cellcolor{white} & P\&R &
\textbf{53.45} & \textbf{95.60} & 0.08 & \textbf{0.49}&\textbf{0.20} & \multicolumn{1}{c|}{\textbf{0.05}} & \textbf{53.33} & \textbf{98.55} &\textbf{0.08} & \textbf{0.00} & \textbf{0.49} & 0.17 \\
\midrule
\multirow{7}{*}{\begin{tabular}[l]{@{}l@{}}+ Initial State \\ \& Object Loc.\end{tabular}}
& WANDR~$\dagger$\cite{diomataris2024wandr}& 33.28 & 80.92 & 0.47 &40.35 &0.42 & \multicolumn{1}{c|}{0.11} & 31.74 & 96.81 & 0.62 & 60.14 & \textbf{0.61} & 0.25 \\
& WANDR~\cite{diomataris2024wandr} & 33.54 & 74.54 & 0.54 & 47.91 & 0.49 & \multicolumn{1}{c|}{0.16} & 49.42 & 98.70 & 0.69 & 68.99 & 0.66 & 0.24\\
& DNO~$\dagger$\cite{karunratanakul2024optimizing} & 37.34 & \textbf{100.00} & 0.47  & 37.87 & 0.44 & \multicolumn{1}{c|}{0.05} & 27.24 & \textbf{100.00}& 0.64 & 59.56 & 0.87 & \textbf{0.09} \\
& DNO~\cite{karunratanakul2024optimizing} & 
46.10 & \textbf{100.00} & 0.42  & 33.33 & 0.29 & \multicolumn{1}{c|}{0.05} & 32.40 & \textbf{100.00} & 0.52 & 45.43 & 0.71 & 0.16 \\
& DartControl~$\dagger$~\cite{Zhao:DartControl:2025} & 28.82 & 89.20 & 0.47 & 40.82 & 0.52 & \multicolumn{1}{c|}{0.13} & 42.90 & \textbf{100.00} & 0.54 & 50.29 & 0.72 & 0.40\\
& DartControl~\cite{Zhao:DartControl:2025} & 30.68 & 89.29 & 0.44 & 35.56 & 0.46 & \multicolumn{1}{c|}{0.08} & 42.03 & \textbf{100.00} & 0.56 & 52.17 & 0.75 & 0.54\\
\rowcolor{LightRed}
\cellcolor{white} & P\&R & \textbf{53.70}& \textbf{100.00} & \textbf{0.37}&\textbf{24.36} & \textbf{0.25}& \multicolumn{1}{c|}{\textbf{0.04}} & \textbf{61.59} & \textbf{100.00} & \textbf{0.51} & \textbf{42.44} & 0.69 & 0.17 \\
\bottomrule
\end{tabular}}
\end{subtable}
\begin{subtable}[t]{\linewidth}
\resizebox{\textwidth}{!}{
\begin{tabular}{cccccccc|cccccc}
\toprule
& & \multicolumn{6}{c|}{HOT3D}& \multicolumn{6}{c}{ADT}\\  
\midrule
Condition& Method & \begin{tabular}[c]{@{}c@{}}Prime \\ Success\end{tabular} $\uparrow$ & \begin{tabular}[c]{@{}c@{}}Reach \\ Success\end{tabular} $\uparrow$ &  \begin{tabular}[c]{@{}c@{}}Goal \\ MPJPE\end{tabular} $\downarrow$  & Loc Err $\downarrow$ & MPJPE $\downarrow$   & \begin{tabular}[c]{@{}c@{}}Foot \\ Skating\end{tabular} $\downarrow$ & \begin{tabular}[c]{@{}c@{}}Prime \\ Success\end{tabular} $\uparrow$ & \begin{tabular}[c]{@{}c@{}}Reach \\ Success\end{tabular} $\uparrow$ & \begin{tabular}[c]{@{}c@{}}Goal \\ MPJPE\end{tabular} $\downarrow$  & Loc Err $\downarrow$ & MPJPE $\downarrow$ & \begin{tabular}[c]{@{}c@{}}Foot \\ Skating\end{tabular} $\downarrow$\\
\midrule
No condition & Static & 0 & 26.43& 0.35 & 22.01 & 0.32 & \multicolumn{1}{c|}{--} & 0 & 9.90 & 1.86 & 69.27 & 1.03 & -- \\
\midrule
\multirow{3}{*}{Text} & MDM $\dagger \ddagger$ & 27.28 & 6.25 & 1.11 & 75.85 & 0.89 & \multicolumn{1}{c|}{0.31} & 9.38 & 5.21 & 2.54 & 97.40 & 1.76 & 0.35 \\
 & MDM $\dagger ^*$ & 8.65 & 4.30 & 0.51 & 36.20 & 0.44 & \multicolumn{1}{c|}{0.00} & 3.65 & 6.25 & 1.98 & 83.33 & 1.16 & 0.06 \\
 & MDM &
 27.86& 34.65 & 0.45 & 18.45 & 0.37& \multicolumn{1}{c|}{0.00} & 11.76 & 17.64 & 1.95 & 89.41 & 1.20 & 0.18 \\
\midrule
\multirow{7}{*}{\begin{tabular}[l]{@{}l@{}}+ Initial State \\ \& Goal Pose\end{tabular}}
& GMD~$\dagger$\cite{karunratanakul2023gmd} & 31.85 & 20.16 & 0.38 & 25.00 & 0.37 & \multicolumn{1}{c|}{0.02} & 16.47 & 12.94 & 0.35 & 10.58 & 0.48 & 0.21 \\
& GMD~\cite{karunratanakul2023gmd} & 
46.84 & 73.39 & 0.21 & 6.87 & 0.34 & \multicolumn{1}{c|}{0.03} & 22.35 & 25.88 & 0.32 & 9.41 & 0.43 & 0.22 \\

& DNO~$\dagger$\cite{karunratanakul2024optimizing} & 45.83 & 90.18 & 0.05 & 9.67 & 0.23 & \multicolumn{1}{c|}{0.02} & 25.88 & 52.94 & \textbf{0.04} & 5.88 & 0.56 & \textbf{0.08} \\
& DNO~\cite{karunratanakul2024optimizing} &
47.10 & 92.11 & \textbf{0.03} & 0.72 & 0.18 & \multicolumn{1}{c|}{0.02} & 28.23 & \textbf{82.35} & 0.05 & \textbf{1.17} & 0.43 & 0.16 \\
& DartControl~$\dagger$~\cite{Zhao:DartControl:2025} & 40.51 & 89.77 & 0.17 & 2.15 & 0.24 & \multicolumn{1}{c|}{0.01} & 14.12 & 67.06 & 0.22 & 8.24 & 0.62 & 0.24 \\
& DartControl~\cite{Zhao:DartControl:2025} & 46.84 & 89.91 & 0.17 & 2.15 & 0.27 & \multicolumn{1}{c|}{\textbf{0.00}} & 23.53 & 56.47 & 0.33 & 10.59 & 0.68 & 0.19\\
\rowcolor{LightRed}
\cellcolor{white} & P\&R & \textbf{53.22}& \textbf{93.81} & 0.06 & \textbf{0.00} & \textbf{0.11} & \multicolumn{1}{c|}{\textbf{0.00}} & \textbf{37.64}& 70.58 & 0.08 & \textbf{1.17} & \textbf{0.33} & 0.11 \\
\midrule
\multirow{7}{*}{\begin{tabular}[l]{@{}l@{}}+ Initial State \\ \& Object Loc.\end{tabular}}
& WANDR~$\dagger$\cite{diomataris2024wandr} & 34.32 & 91.66 & 0.33 & 17.77 & 0.27 & \multicolumn{1}{c|}{0.05} & 7.65 & 82.94 & 0.66 & 61.76 & 0.60 & 0.22\\ 
& WANDR~\cite{diomataris2024wandr} & 41.45 & 91.66 & 0.33 & 17.50 & 0.24 & \multicolumn{1}{c|}{0.04} & 38.24 & 80.00 & 0.66 & 58.24 & 0.57 & 0.21\\
& DNO~$\dagger$~\cite{karunratanakul2024optimizing} & 46.77 & \textbf{100.00} & 0.41 & 40.86 & 0.36 & \multicolumn{1}{c|}{0.04} & 23.53 & \textbf{100.00} & 0.64 & 61.18 & 0.70 & \textbf{0.10} \\
& DNO~\cite{karunratanakul2024optimizing}& 
58.06 & \textbf{100.00} & 0.35 & 20.16& 0.30 & 0.01& 38.82 & \textbf{100.00} & 0.60 & 57.79 & 0.59 & \textbf{0.10} \\
& DartControl~$\dagger$~\cite{Zhao:DartControl:2025} & 46.43 & 97.58 & 0.38 & 19.78 & 0.34 & \multicolumn{1}{c|}{0.01} & 25.88 & 97.65 & 0.69 & 64.71 & 0.86 & 0.29 \\
& DartControl~\cite{Zhao:DartControl:2025} & 45.09 & 96.64 & 0.33 & 9.15 & 0.30 & \multicolumn{1}{c|}{\textbf{0.00}} & 31.76 & 96.47 & 0.69 & 72.94 & 0.75 & 0.21\\
\rowcolor{LightRed}
\cellcolor{white} & P\&R & \textbf{65.45}& \textbf{100.00} & \textbf{0.25} & \textbf{4.16} & \textbf{0.16} & \multicolumn{1}{c|}{0.01} & \textbf{43.53}& \textbf{100.00} &\textbf{0.55} & \textbf{52.54} & \textbf{0.48} & 0.14\\

\bottomrule 
\end{tabular}}
\end{subtable}

\begin{minipage}[t]{0.59\linewidth}
\resizebox{\textwidth}{!}{
\begin{tabular}{cccccccc}
\toprule
& & \multicolumn{6}{c}{GIMO}\\  
\midrule
Condition& Method & \begin{tabular}[c]{@{}c@{}}Prime \\ Success\end{tabular} $\uparrow$ & \begin{tabular}[c]{@{}c@{}}Reach \\ Success\end{tabular} $\uparrow$ & \begin{tabular}[c]{@{}c@{}}Goal \\ MPJPE\end{tabular} $\downarrow$  & Loc Err $\downarrow$ & MPJPE $\downarrow$   & \begin{tabular}[c]{@{}c@{}}Foot \\ Skating\end{tabular} $\downarrow$ \\
\midrule
No condition & Static & 0 & 0 & 3.41 & 100.0 & 1.86 & -- \\
\midrule
\multirow{3}{*}{Text} & MDM $\dagger \ddagger$ & 0 & 0 & 3.69 & 100 & 2.17 & 0.45 \\
\multicolumn{1}{c}{} & MDM $\dagger ^*$ & 0  & 0 & 3.34 & 100 & 1.82 & 0.08 \\
\multicolumn{1}{c}{} & MDM & 
0 & 0 & 2.74 & 100 & 1.56 & 0.12 \\ 
\midrule
\multirow{7}{*}{\begin{tabular}[l]{@{}l@{}}+ Initial State \\ \& Goal Pose\end{tabular}}
& GMD~$\dagger$\cite{karunratanakul2023gmd} & 13.63 & 4.76 & 0.44 & 11.90 & 0.62 & 0.43 \\
& GMD~\cite{karunratanakul2023gmd} & 22.72 &9.09 & 0.34& 9.09& 0.58& 0.19 \\
& DNO~$\dagger$~\cite{karunratanakul2024optimizing} & 18.18 & 27.27 & 0.15 & 9.09 & 0.80 & 0.18 \\
& DNO~\cite{karunratanakul2024optimizing} & 31.82& 36.36& 0.16& \textbf{4.54}&0.62& 0.18 \\
& DartControl~$\dagger$~\cite{Zhao:DartControl:2025} & 9.52 & 14.29 & 0.33 & 9.52 & 1.37 & 0.07 \\
& DartControl~\cite{Zhao:DartControl:2025} & 38.10 & 9.52 & 0.35 & 9.52 & 0.85 & \textbf{0.04}\\
\rowcolor{LightRed}
\cellcolor{white} & P\&R & \textbf{45.45}& \textbf{59.09} & \textbf{0.13} & \textbf{4.54} & \textbf{0.55} & 0.18 \\
\midrule
\multirow{7}{*}{\begin{tabular}[l]{@{}l@{}}+ Initial State \\ \& Object Loc.\end{tabular}}
& WANDR~$\dagger$\cite{diomataris2024wandr} & 14.29 & 61.90 & 0.51 & 57.14 & 0.79 & 0.44 \\ 
& WANDR~\cite{diomataris2024wandr} & 14.29 & 66.67 & 0.51 & 47.62 & \textbf{0.57} & 0.43\\
& DNO~$\dagger$~\cite{karunratanakul2024optimizing} & 27.27 & \textbf{100.00} & 0.76 & 63.63 & 0.98 & 0.11 \\
& DNO~\cite{karunratanakul2024optimizing} & 
36.36 & \textbf{100.00} & 0.45 & 50.00 & 0.70 & 0.12 \\
& DartControl~$\dagger$~\cite{Zhao:DartControl:2025} & 4.76 & 76.19 & 0.49 & 61.90 & 1.51 & 0.10\\
& DartControl~\cite{Zhao:DartControl:2025} & \textbf{57.14} & 66.67 & 0.44 & 47.62 & 0.87 & \textbf{0.05}\\
\rowcolor{LightRed}
\cellcolor{white} & P\&R & 50.00 & \textbf{100.00} & \textbf{0.38} & \textbf{36.36} & 0.63 & 0.18 \\
\bottomrule 
\end{tabular}}
\end{minipage}
\end{table}

\section{Ablation of Architecture \& Loss}

We present additional ablations of our architecture and losses. As in the main paper, we train a single model on all datasets, and report results on two datasets: HD-EPIC and MoGaze.

\subsection{Transformer Encoder v/s Decoder} We compare performance of transformer encoder v/s decoder based diffusion model for the task of \pr\ motion generation in \cref{tab:encoder_vs_decoder}. While the decoder architecture injects condition $\mathbf{\hat{z}_t}$ by cross-attention with each decoder layer, the encoder provides the condition as an additional token at the input of the first encoder layer. The decoder architecture performs significantly better than the encoder architecture, making it a superior choice for the task.

\begin{table*}[t!]
\centering
\caption{\textbf{Encoder v/s Decoder}. We compare encoder and decoder architecture for \pr\ motion generation.}
\label{tab:encoder_vs_decoder}
\vspace*{-8pt}
\resizebox{\linewidth}{!}{
\begin{tabular}{lcccc|cccc}
\toprule
& \multicolumn{4}{c|}{HD-EPIC} & \multicolumn{4}{c}{MoGaze}\\
\midrule
Architecture & \begin{tabular}[c]{@{}c@{}}Prime \\ Success\end{tabular} $\uparrow$ & \begin{tabular}[c]{@{}c@{}}Reach \\ Success\end{tabular} $\uparrow$   & Loc Err $\downarrow$ & MPJPE $\downarrow$ & \begin{tabular}[c]{@{}c@{}}Prime \\ Success\end{tabular} $\uparrow$ & \begin{tabular}[c]{@{}c@{}}Reach \\ Success\end{tabular} $\uparrow$   & Loc Err $\downarrow$ & MPJPE $\downarrow$ \\
\midrule
Encoder &35.18 &94.05&56.19&0.54& 56.80 & 90.89 & 59.29 & 0.85 \\
\rowcolor{LightRed}Decoder & \textbf{51.00}&\textbf{100.00}&\textbf{25.26}&\textbf{0.27}& \textbf{62.37} & \textbf{100.00} & \textbf{36.43} & \textbf{0.56} \\
\bottomrule 
\end{tabular}
}
\end{table*}

\subsection{Training Loss}
We ablate the impact of $\mathcal{L}_{\text{joint}}$ in \cref{tab:loss_ablation}. We find adding the $\mathcal{L}_{\text{joint}}$ helps improve \pr\ generation for both HD-EPIC and MoGaze.

\begin{table*}[t!]
\centering
\caption{\textbf{Loss Ablation}.
}

\label{tab:loss_ablation}
\vspace*{-8pt}
\resizebox{\linewidth}{!}{
\begin{tabular}{lcccc|cccc}
\toprule
& \multicolumn{4}{c|}{HD-EPIC} & \multicolumn{4}{c}{MoGaze}\\
\midrule
Loss & \begin{tabular}[c]{@{}c@{}}Prime \\ Success\end{tabular} $\uparrow$ & \begin{tabular}[c]{@{}c@{}}Reach \\ Success\end{tabular} $\uparrow$   & Loc Err $\downarrow$ & MPJPE $\downarrow$ & \begin{tabular}[c]{@{}c@{}}Prime \\ Success\end{tabular} $\uparrow$ & \begin{tabular}[c]{@{}c@{}}Reach \\ Success\end{tabular} $\uparrow$   & Loc Err $\downarrow$ & MPJPE $\downarrow$ \\
\midrule
$\mathcal{L}$& \textbf{51.65}&\textbf{100.00}&33.22&0.34& 61.10 & \textbf{100.00} & 57.24 & 0.70 \\
\rowcolor{LightRed}$\mathcal{L}+\mathcal{L}_{\text{joint}}$ & 51.00&\textbf{100.00}&\textbf{25.26}&\textbf{0.27}& \textbf{62.37} & \textbf{100.00} & \textbf{36.43} & \textbf{0.56} \\
\bottomrule 
\end{tabular}
}
\end{table*}

\subsection{Incorporating Goal Condition}
We condition our \pr\ model by adding the initial state and goal pose/target location condition $\mathbf{p}$ to the text condition $\mathbf{z_t}$ to get $\mathbf{\hat{z}}_t$.
We ablate another alternative of incorporating $\mathbf{p}$ to $\mathbf{z_t}$ using cross-attention as shown in 
\begin{gather}
    \mathbf{\delta_t} = CA(\mathbf{z_t}, \mathbf{p}) = \text{Softmax}\left(\frac{(\mathbf{z_t} \mathbf{W}_Q) (\mathbf{p} \mathbf{W}_K)^T)}{\sqrt{d_k}}\right) (\mathbf{p} \mathbf{W}_V) \nonumber \\
    \mathbf{\hat{z}_t} = \mathbf{z_t} + \mathbf{\delta_t}
\end{gather}
where $CA$ is a 1-layer cross-attention. $\mathbf{z_t}$ is linearly projected to get the query and $\mathbf{p}$ is projected to key and value. 
We use a residual network to make the most of our pretraining. We show the results in \cref{tab:condition_injection}. We find that incorporating the condition through addition performs better, across 7 out of the 8 metrics.

\begin{table*}[t!]
\centering
\caption{\textbf{Condition Injection}. We verify different methods for injecting our initial state and goal conditions.}
\label{tab:condition_injection}
\resizebox{\linewidth}{!}{
\begin{tabular}{lcccc|cccc}
\toprule
& \multicolumn{4}{c|}{HD-EPIC} & \multicolumn{4}{c}{MoGaze}\\
\midrule
Method & \begin{tabular}[c]{@{}c@{}}Prime \\ Success\end{tabular} $\uparrow$ & \begin{tabular}[c]{@{}c@{}}Reach \\ Success\end{tabular} $\uparrow$   & Loc Err $\downarrow$ & MPJPE $\downarrow$ & \begin{tabular}[c]{@{}c@{}}Prime \\ Success\end{tabular} $\uparrow$ & \begin{tabular}[c]{@{}c@{}}Reach \\ Success\end{tabular} $\uparrow$   & Loc Err $\downarrow$ & MPJPE $\downarrow$ \\
\midrule
$CA$ & 50.20&\textbf{100.00}&33.45&0.36& \textbf{65.30} & \textbf{100.00} & 49.34 & 0.68 \\
\rowcolor{LightRed}Addition & \textbf{51.00}&\textbf{100.00}&\textbf{25.26}&\textbf{0.27}& 62.37 & \textbf{100.00} & \textbf{36.43} & \textbf{0.56} \\
\bottomrule 
\end{tabular}
}
\end{table*}

\subsection{Ablation of Number of Diffusion Steps $T$}
For a single motion generation, the diffusion model starts from noise at $t=T$ and iteratively denoises it over diffusion steps $t=\{T, T-1, \cdots, 0\}$, finally producing the clean motion at $t=0$. We ablate the choice of $T$ in \cref{tab:ablation_of_T}, which controls the total number of steps needed to generate a sequence of motion. We find that $T=50$ gives consistently good performance across all metrics, with at least a $+1.30\%$ improvement in prime success on HD-EPIC.
Importantly, the method is generally robust to the number of steps.

\begin{table*}[t!]
\centering
\caption{\textbf{Impact of diffusion steps $T$}. We compare the performance of \pr\ motion generation for multiple diffusion steps.}
\label{tab:ablation_of_T}
\vspace*{-8pt}
\resizebox{\linewidth}{!}{
\begin{tabular}{lcccc|cccc}
\toprule
& \multicolumn{4}{c|}{HD-EPIC} & \multicolumn{4}{c}{MoGaze}\\
\midrule
$T$ & \begin{tabular}[c]{@{}c@{}}Prime \\ Success\end{tabular} $\uparrow$ & \begin{tabular}[c]{@{}c@{}}Reach \\ Success\end{tabular} $\uparrow$   & Loc Err $\downarrow$ & MPJPE $\downarrow$ & \begin{tabular}[c]{@{}c@{}}Prime \\ Success\end{tabular} $\uparrow$ & \begin{tabular}[c]{@{}c@{}}Reach \\ Success\end{tabular} $\uparrow$   & Loc Err $\downarrow$ & MPJPE $\downarrow$ \\
\midrule
10 & 49.45& \textbf{100.00}&28.46&0.30& \textbf{63.50} & \textbf{100.00} & 54.02 & 0.64 \\
\rowcolor{LightRed}50 & \textbf{51.00}&\textbf{100.00}&\textbf{25.26}&\textbf{0.27}& 62.37 & \textbf{100.00} & \textbf{36.43} & \textbf{0.56} \\
100 & 50.30&\textbf{100.00}&25.76&0.28& 62.00 & \textbf{100.00} & 48.86 & 0.61 \\
500 &49.80 & \textbf{100.00}&27.46&0.31& 61.00 & \textbf{100.00} & 49.95 & 0.65\\
1000 & 49.65&\textbf{100.00}&27.01&0.33 &61.45 & \textbf{100.00} & 51.10 & 0.68\\
\bottomrule 
\end{tabular}
}
\vspace*{-12pt}
\end{table*}

\begin{figure*}[t]
    \centering
    \includegraphics[width=\linewidth]{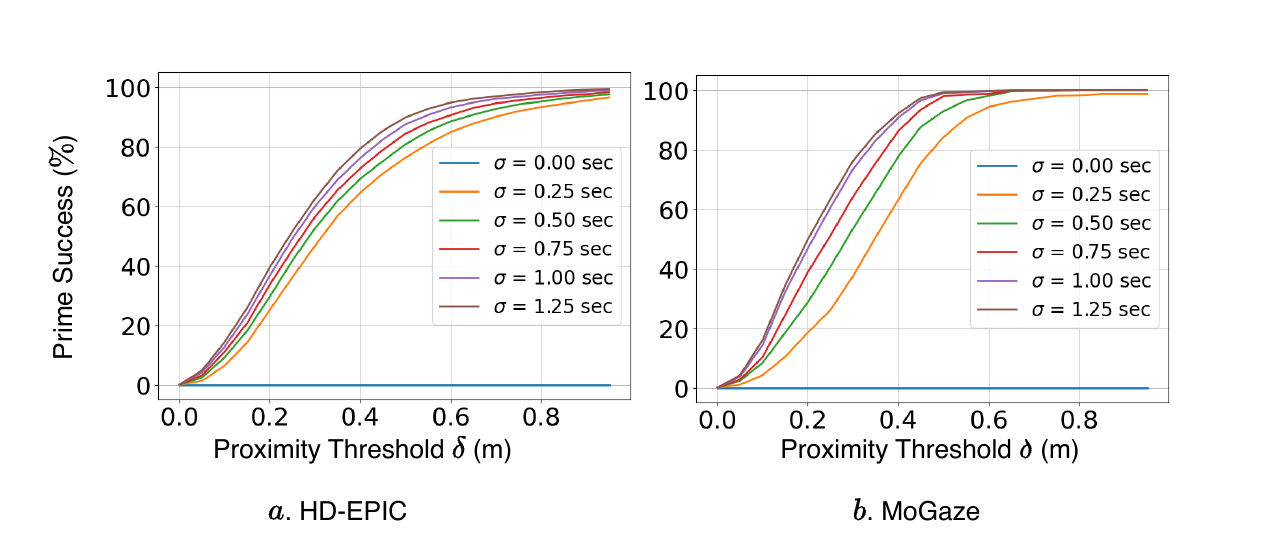}
    \caption{Varying time window $\sigma$ and proximity threshold $\delta$ for Prime Success calculation on HD-EPIC and MoGaze.}
    \label{fig:sigma_theta_analysis}
\end{figure*}

\section{Prime Success Metric: Analysis}
\label{sec:sigma_theta_analysis}
We conduct an in-depth analysis to better understand the impact of hyperparameters (the time window $\sigma$ and proximity threshold $\delta$ for determining gaze-object intersection) used in calculating the newly introduced Prime Success metric.
We compare our \pr\ predictions 
while varying the hyperparameters of the metric.
Note that as the thresholds are changed, the motion can be considered a success or a failure.
Recall that our results are reported for $\delta=0.25 m$ and $\sigma=1.0$ sec. 

To evaluate the impact of these hyperparameters, we vary $\delta$ on the x-axis (between 0 and  1 m), then plot distinct curves for discrete time windows: 0, 0.25, 0.5, 0.75, 1.0, 1.25 seconds. \cref{fig:sigma_theta_analysis} shows that a very tight time window is too restrictive for MoGaze. As expected, a high $\delta$ threshold is too permissive and cannot be used to compare different methods.

\section{Evaluating Pre-trained Models}

In Tab. 4 in the main manuscript, we presented results showcasing the impact of the pre-training dataset on our model's performance.
For completion, we here also present results on evaluating the pre-trained models themselves, before we do any fine-tuning.
We evaluate the models trained on HumanML3D and Nymeria for the task of text-conditioned motion generation. 

As in previous works~\cite{tevet2023mdm}, we train motion-text embedding models~\cite{Guo2022t2m}, with two encoders: one for motion and one for text, using a contrastive loss. We use paired text-motion sequences from the Nymeria train set. We follow the architecture for our encoders from ~\cite{Guo2022t2m}. 

Following \cite{guo2020action2motion,tevet2023mdm}, we use the following metrics for evaluation - 
\begin{itemize}
    \item R Precision (Top-3): Given batches of motion and corresponding text, the most similar texts to each motion are ranked based on the Euclidean distances. This calculates the percentage of motion sequences for which the correct text is retrieved in the top 3 matches.
    \item FID: This compares the encoded feature distribution of the generated motion to that of real motion
    \item Multimodal Distance: This calculates the average Euclidean distance in the embedding space between paired motion and text.
    \item Diversity: This measures the variance in the generated motion over all text prompts in the test set.
\end{itemize}

\begin{figure*}[!t]
    \centering
\includegraphics[width=\linewidth]{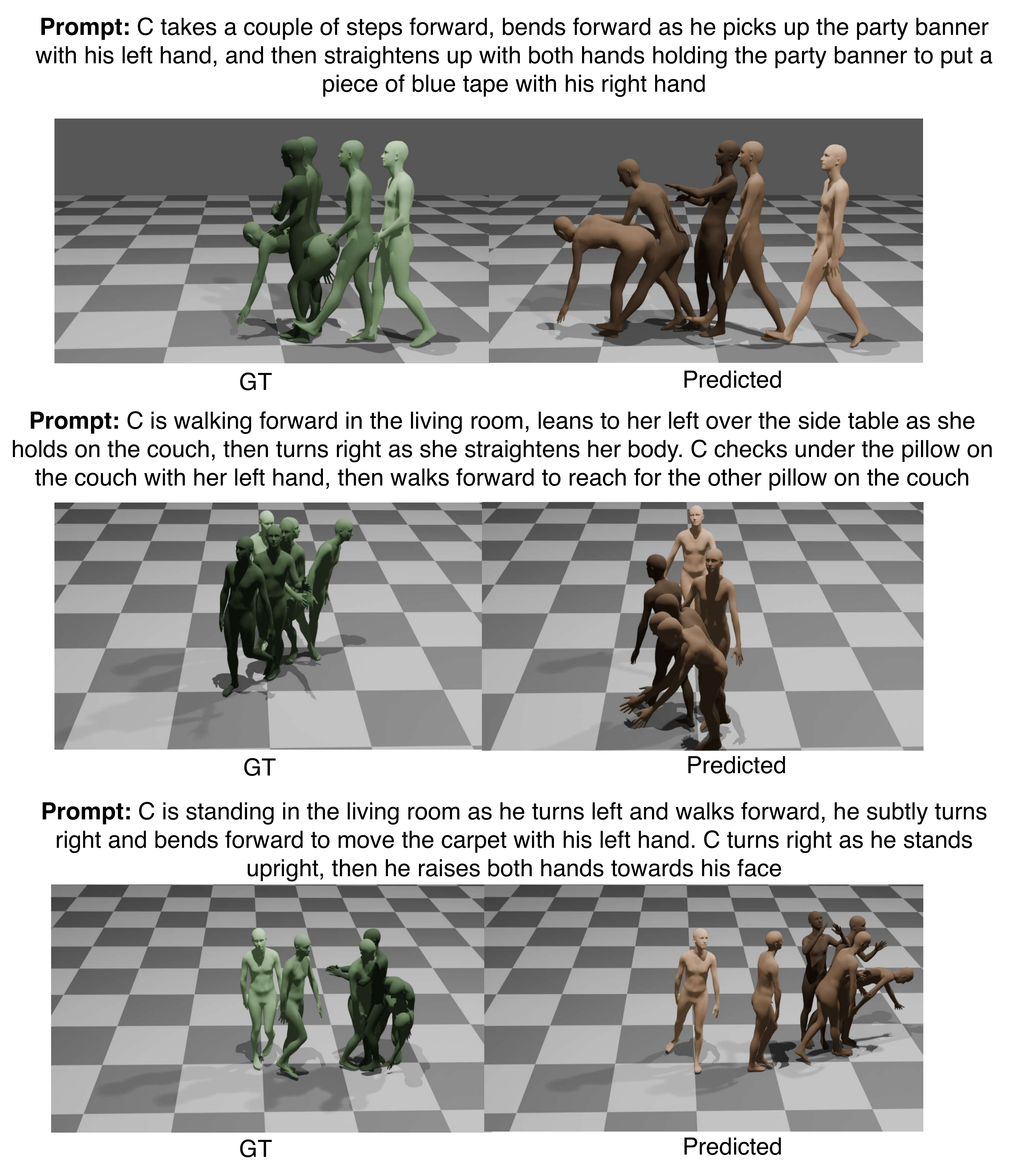}
\caption{Qualitatives of our pre-trained model. We showcase both ground truth (GT) and predicted motion for given text prompts. Darker poses represent later times.}
\label{fig:pretraining_Examples}
\end{figure*} 

We provide the results of our pretrained model in \cref{tab:pretraining_results}. We find that the motion generated by our Nymeria-pretrained model aligns better with the fine-grained texts of Nymeria. This is verified by the $+34.0\%$ and $-2.68$ improvements in R-Precision and Multi-modal distance respectively. The diversity of our generated motions is $+2.49$ higher than that of motions generated by the HumanML3D pre-trained model. We showcase some of the qualitative results of the pre-trained model in \cref{fig:pretraining_Examples}.

\begin{table}[t!]
\centering
\caption{\textbf{Pretraining results}. }

\label{tab:pretraining_results}
\vspace*{-8pt}
\resizebox{\linewidth}{!}{
\begin{tabular}{lccccc}
\toprule
\begin{tabular}[c]{@{}c@{}}Pretraining \\Dataset\end{tabular} & Motion & \begin{tabular}[c]{@{}c@{}}R Precision \\ (Top- 3)\end{tabular}$\uparrow$ & FID $\downarrow$& \begin{tabular}[c]{@{}c@{}}Multi-modal\\ Distance\end{tabular} $\downarrow$ & Diversity $\uparrow$  \\
\midrule
\rowcolor{LightGrey}& Real&75.43 $\pm$ 0.12& 0 &2.79$\pm$ 0.00   & 9.70 $\pm$ 0.13\\
HumanML3D& Generated&43.32 $\pm$ 0.69& 11.39 $\pm$ 0.64 &5.53$\pm$ 0.08   & 7.26 $\pm$ 0.08\\

\rowcolor{LightRed} Nymeria& Generated&\textbf{77.25 $\pm$ 0.42}& \textbf{0.97 $\pm$ 0.11} &\textbf{2.85$\pm$ 0.03}   & \textbf{9.75 $\pm$ 0.09}\\
\bottomrule 
\end{tabular}}
\end{table}

\end{document}